\documentclass[10pt]{article} 
\usepackage[preprint]{tmlr}

\usepackage{hyperref}
\usepackage{url}
\usepackage[utf8]{inputenc} 
\usepackage[T1]{fontenc}    
\usepackage{booktabs}       
\usepackage{amsfonts}       
\usepackage{nicefrac}       
\usepackage{microtype}      
\usepackage{xcolor}         

\usepackage{graphicx} 
\usepackage{amsmath}
\usepackage{amssymb}
\usepackage{mathtools}
\usepackage{multirow}
\usepackage{caption}
\usepackage{color}
\usepackage{dsfont} 
\usepackage[normalem]{ulem} 

\newcommand{\textb}[1]{\textcolor{black}{#1}}

\renewcommand{\tilde}{\widetilde}
\renewcommand{\hat}{\widehat}

\newcommand{\defn}{\triangleq}

\newcommand{\mc}[1]{\ensuremath{\mathcal{#1}}}

\newcommand{\Real}{{\mathbb{R}}}

\newcommand{\of}[1]{^{{(#1)}}}

\newcommand{\tran}{^{\top}}

\DeclareMathOperator{\E}{E}

\DeclareMathOperator{\EmpQuant}{EmpQuant}

\newcommand{\test}{_{\mathsf{test}}}
\renewcommand{\cal}{_{\mathsf{cal}}}
\newcommand{\train}{_{\mathsf{train}}}

\renewcommand{\eqref}[1]{(\ref{eq:#1})}
\newcommand{\secref}[1]{Sec.~\ref{sec:#1}}
\newcommand{\Secref}[1]{Section~\ref{sec:#1}}
\newcommand{\figref}[1]{Fig.~\ref{fig:#1}}
\newcommand{\Figref}[1]{Figure~\ref{fig:#1}}
\newcommand{\tabref}[1]{Tab.~\ref{tab:#1}}
\newcommand{\Tabref}[1]{Table~\ref{tab:#1}}
\newcommand{\appref}[1]{App.~\ref{app:#1}}
\newcommand{\Appref}[1]{Appendix~\ref{app:#1}}

\DeclareMathOperator{\EC}{EC}

\title{Conformal Bounds on Full-Reference Image Quality for Imaging Inverse Problems}

\author{\name Jeffrey Wen \email wen.254@osu.edu \\
      \addr Department of Electrical and Computer Engineering\\
      The Ohio State University
      \AND
      \name Rizwan Ahmad \email rizwan.ahmad@osumc.edu \\
      \addr Department of Biomedical Engineering\\
      The Ohio State University
      \AND
      \name Philip Schniter \email schniter.1@osu.edu\\
      \addr Department of Electrical and Computer Engineering\\
      The Ohio State University
}

\def\month{05}  
\def\year{2025} 
\def\openreview{\url{https://openreview.net/forum?id=WADLPccB6o}} 

\begin{document}

\maketitle

\begin{abstract}
In imaging inverse problems, we would like to know how close the recovered image is to the true image in terms of full-reference image quality (FRIQ) metrics like PSNR, SSIM, LPIPS, etc.
This is especially important in safety-critical applications like medical imaging, where knowing that, say, the SSIM was poor could potentially avoid a costly misdiagnosis.
But since we don't know the true image, computing FRIQ is non-trivial.
In this work, we combine conformal prediction with approximate posterior sampling to construct bounds on FRIQ that are guaranteed to hold up to a user-specified error probability.
We demonstrate our approach on image denoising and accelerated magnetic resonance imaging (MRI) problems.
Code is available at \url{https://github.com/jwen307/quality_uq}.
\end{abstract}

\section{Introduction}
\label{sec:introduction}

In imaging inverse problems, one aims to recover a true image $x_0$ from noisy/distorted/incomplete measurements $y_0=\mc{A}(x_0)$ \citep{Arridge:AN:19}. 
Denoising, deblurring, inpainting, super-resolution, limited-angle computed tomography, and accelerated magnetic resonance imaging (MRI) are examples of linear inverse problems, while phase-retrieval, de-quantization, low-light imaging, and image-to-image translation are examples of non-linear inverse problems. 
Such problems are ill-posed, in that many hypotheses of $x_0$ are consistent with both the measurements $y_0$ and prior knowledge about $x_0$.
To complicate matters, different recovery methods are biased towards different plausible image hypotheses, leading to important differences in reconstruction quality.
For example, modern deep-network approaches can sometimes hallucinate 
\citep{
Cohen:MICCAI:18,%
Belthangady:NMe:19,%
Hoffman:NMe:21,%
Muckley:TMI:21,%
Bhadra:TMI:21,%
Gottschling:23,%
Tivnan:MICCAI:24}, 
i.e., generate visually pleasing recoveries that differ in important ways from the true image $x_0$. 
Thus, there is a strong need to quantify the accuracy of a given recovery, especially in safety-critical applications like medical imaging \citep{Chu:JACR:20,Banerji:NM:23}.

In image recovery, ``accuracy'' can be defined in different ways.
Classical metrics like mean-squared error (MSE), or its scaled counterpart peak signal-to-noise ratio (PSNR), are convenient for theoretical analysis but do not always correlate well with human perceptions of image quality. 
This fact inspired the field of full-reference image-quality (FRIQ) assessment \citep{Lin:JVCIR:11,Wang:SPM:11}, which led to the well-known Structural Similarity Index Measure (SSIM) \citep{Wang:TIP:04} that is still popular today.
However, progress continues to be made.
Most recent methods leverage the internal features of deep neural networks, which are said to mimic the processing architecture of the human visual cortex \citep{Yamins:NNS:16,Lindsay:JCNS:21}.
A popular example of the latter is Learned Perceptual Image Patch Similarity (LPIPS) \citep{Zhang:CVPR:18}.
In the end, though, the best choice of metric may depend on the application.
For example, in magnetic resonance imaging (MRI), the goal is to provide the radiologist with an image recovery that leads to an accurate diagnosis.
A recent clinical MRI study \citep{Kastryulin:IA:23} found that, among 35 tested metrics, Deep Image Structure and Texture Similarity (DISTS) \citep{Ding:TPAMI:20} correlated best with radiologists' ratings \textb{of perceived noise level, contrast level, and presence of artifacts when comparing reconstructions to the fully sampled ground-truth}. 

In this work, our goal is to provide rigorous bounds on the FRIQ $m(\hat{x}_0,x_0)$ of a recovery $\hat{x}_0=h(y_0)$ relative to the true image $x_0$.
Here, $h(\cdot)$ is an arbitrary image-recovery scheme and $m(\cdot,\cdot)$ is an arbitrary FRIQ metric.
The key challenge is that $x_0$ is unknown.
To our knowledge, there exists no prior work on providing FRIQ guarantees in image recovery.
Our contributions are as follows.
\begin{enumerate}
\item 
We propose a framework to bound the FRIQ $m(\hat{x}_0,x_0)$ of a recovered image $\hat{x}_0$ without access to the true image $x_0$.  Our framework uses conformal prediction \citep{Vovk:Book:05, Angelopoulos:FTML:23} to construct bounds that hold with probability at least $1-\alpha$ under certain exchangeability assumptions and where $\alpha\in (0,1)$ is chosen by the user.
\item 
We show how posterior-sampling-based image recovery can be used to construct conformal bounds that adapt to the measurements $y_0$ and reconstruction $\hat{x}_0$.
\item 
We demonstrate our approach on two linear inverse problems: denoising of FFHQ faces \citep{Karras:CVPR:19} faces and recovery of fastMRI knee images \citep{Zbontar:18} from accelerated multicoil measurements.
\end{enumerate}

From the perspective of uncertainty quantification (UQ), one could say that our goal is to bound the uncertainty on FRIQ $m(\hat{x}_0,x_0)$ that arises due to $x_0$ being unknown.
As such, our approach to UQ differs from typical ones in image recovery. 
There, uncertainty is typically quantified on individual pixels, with the overall result being a pixel-wise uncertainty map.
To construct these maps, it's popular to use (approximate) posterior samplers
\citep{Adler:18,Tonolini:JMLR:20,Edupuganti:TMI:20,Jalal:NIPS:21,Sun:AAAI:21,Laumont:JIS:22,Zach:TMI:23,Wen:ICML:23,Bendel:NIPS:23,Wu:24}
or Bayesian neural networks (BNNs) 
\citep{Kendall:NIPS:17, Xue:Optica:19, Barbano:ICPR:21, Ekmekci:TCI:22, Narnhofer:TMI:22}
to draw many reconstructions from the distribution of plausible $x_0$ for a given $y_0$ (i.e., the posterior distribution $p_{X_0|Y_0}(\cdot|y_0)$), 
from which pixel-wise standard-deviations can be estimated. 
An alternative is to utilize conformal prediction to produce pixel-wise intervals that are guaranteed to contain the true pixel value with high probability \citep{Angelopoulos:ICML:22, Horwitz:22, Teneggi:ICML:23, Kutiel:ICLR:23, Narnhofer:JIS:24}.
Although these uncertainty maps can be visually interesting, they do not quantify uncertainty on multi-pixel structures of interest, such as hallucinations or anatomical features relevant to medical diagnosis (e.g., tumors). 

To our knowledge, there exist relatively few works on multi-pixel UQ, and none target FRIQ.
For example, 
\citet{Durmus:JIS:18} use hypothesis testing to infer the presence/absence of a structure-of-interest within the maximum a posteriori (MAP) image recovery, but their method relies on inpainting to construct the structure-absent hypothesis, which may not be accurate.
\citet{Sankaranarayanan:NIPS:22} use conformal prediction to compute uncertainty intervals on the presence/absence of semantic attributes (e.g., whether a face has a smile, glasses, etc.) but their method requires a ``disentangled'' generative adversarial network (GAN) that generates image samples given a set of attribute probabilities.
\citet{Belhasin:TPAMI:23} compute conformal prediction intervals on the principal components of the posterior covariance matrix.
Lastly, given measurements $y_0=\mc{A}(x_0)$ and a downstream imaging task $\mu(\cdot)\in\Real$ (e.g., soft-output classification), \citet{Wen:24} compute conformal bounds on the true task output $\mu(x_0)$. 
While interesting, none of the above works quantify the uncertainty on FRIQ metrics like PSNR, SSIM, LPIPS, DISTS, etc., due to $x_0$ being unknown.
\textb{As metrics like SSIM, DISTS, and LPIPS are designed to quantify higher-level perceptual similarities, bounds on these metrics can provide insight into the differences between multi-pixel structures within the reconstruction and the true image.
For applications like accelerated MRI, where DISTS has been shown to correlate with radiologists' perceptions \citep{Kastryulin:IA:23}, rigorous bounds on FRIQ can provide useful information on the clinical utility of an accelerated reconstruction.}

\section{Background}
\label{sec:background}

Conformal prediction (CP) \citep{Vovk:Book:05, Angelopoulos:FTML:23} is a powerful framework for computing uncertainty intervals on the output of any black-box predictor. 
CP makes no assumptions on the distribution of the data, yet provides probabilistic guarantees that the true target lies within the constructed uncertainty interval. 
In this paper, we focus on the common variant known as split CP \citep{Papadopoulos:ECML:02,Lei:JASA:18}.

%
We now provide a brief background on split CP. Given features $u_0\in\mc{U}$, the goal of CP is to construct a set $\mc{C}_\lambda(\hat{z}_0)$ 
that contains an unknown target $z_0\in\mc{Z}$ with high probability.
\textb{
Here, $\hat{z}_0=f(u_0)$ is some prediction from a black-box model $f(\cdot)$, and $\mc{C}_\lambda(\cdot)$ is constructed so that $\mc{C}_{\lambda}(\hat{z}_0)\subset \mc{C}_{\lambda'}(\hat{z}_0)$ for any $\lambda < \lambda'$ and $\hat{z}_0$.
}
Split CP accomplishes this goal by calibrating $\lambda$ using a dataset of feature and target pairs $\{(u_i,z_i)\}_{i=1}^n$ that has not been used to train $f(\cdot)$.
In particular, it first constructs the set $d\cal \defn \{(\hat{z}_i,z_i)\}_{i=1}^n$ using $\hat{z}_i=f(u_i)$ and then finds a $\hat{\lambda}(d\cal)$ to provide the marginal coverage guarantee \citep{Lei:JRSS:14}
\begin{equation}
\Pr \big\{ Z_{0} \in \mc{C}_{\hat{\lambda}(D\cal)}(\hat{Z}_{0}) \big\} \geq 1-\alpha 
\label{eq:coverage} ,
\end{equation}
where $\alpha$ is a user-chosen error rate.
Here and in the sequel, we use capital letters to denote random variables and lower-case letters to denote their realizations.
In words, \eqref{coverage} guarantees that the unknown target $Z_0$ falls within the interval $\mc{C}_{\hat{\lambda}(D\cal)}(\hat{Z}_{0})$ with probability at least $1-\alpha$ when averaged over the randomness in the test data $(Z_0,\hat{Z}_0)$ and calibration data $D\cal$. 

While there are a number of ways to describe CP calibration of $\lambda$ \citep{Vovk:Book:05, Angelopoulos:FTML:23},
we will focus on the method from \citet{Angelopoulos:22}.
It starts by defining the empirical miscoverage as
\begin{align}
\hat{r}_n(\lambda; d\cal) 
\defn \frac{1}{n} \sum_{i=1}^{n} \mathds{1}\{z_i \notin \mc{C}_{\lambda}(\hat{z}_i)\} 
\label{eq:empirical_miscoverage},
\end{align}
where $\mathds{1}\{\cdot\}$ is the indicator function. 
The empirical miscoverage measures the proportion of targets $z_i$ that land outside of $\mc{C}_{\lambda}(\hat{z}_i)$ in the calibration set $d\cal$.
The calibration procedure then sets $\lambda$ at
\begin{equation}
\hat{\lambda}(d\cal) = \inf \big\{ \lambda : \hat{r}_n(\lambda; d\cal) \leq \alpha - \tfrac{1-\alpha}{n} \big\}
\label{eq:lambda_hat},
\end{equation}
which can be found using a simple binary search. 
Intuitively, the $\lambda$ chosen in \eqref{lambda_hat} yields an empirical miscoverage that is slightly more conservative than the desired $\alpha$ in order to handle the finite size of the calibration set.
When $\{(Z_0,\hat{Z}_0),(Z_1,\hat{Z}_1), \dots,(Z_n,\hat{Z}_n)\}$ are statistically exchangeable, \eqref{lambda_hat} ensures that \eqref{coverage} holds \citep{Angelopoulos:22}.
See the overviews \citep{Angelopoulos:FTML:23, Vovk:Book:05} for more on conformal prediction.

\textb{
For exchangeability, it suffices that the feature/target pairs $(U_i,Z_i)$ are i.i.d across test and calibration samples $i$, although independence is not strictly necessary.
However, when there is distribution shift between the test and calibration samples, exchangeability will be lost.
Works like \citet{Tibshirani:NIPS:19,Barber:AS:23,Cauchois:JASA:24} propose methods to handle distribution shift.
We discuss distribution shift further in \secref{exchange}.
}

\section{Proposed approach} \label{sec:methods}
Consider an imaging inverse problem, where we observe distorted, incomplete, and/or noisy measurements $y_0=\mc{A}(x_0)$ of a true image $x_0$. 
Suppose that $\hat{x}_0=h(y_0)$ is a reconstruction of $x_0$ provided by some image recovery method $h(\cdot)$ and that $z_0 = m(\hat{x}_0,x_0)\in\Real$ is some FRIQ metric on $\hat{x}_0$ with respect to the true $x_0$.
We would like to know $z_0$, especially in safety critical applications.
For example, if $z_0$ was unacceptable, then perhaps we could use a different recovery method $h(\cdot)$ or collect more measurements $y_0$.
But $z_0$ cannot be directly computed because, in practice, $x_0$ is unknown.


Our key insight is that it's possible to construct a set $\mc{C}_\lambda(\hat{z}_0)$ that is guaranteed to contain the unknown FRIQ $z_0$ with high probability.
This can be done using CP, at least when one has access to calibration data $\{(x_i,y_i)\}_{i=1}^n$ 
of true image and measurement pairs that agrees with the test $(x_0,y_0)$ in the sense that the resulting FRIQ pairs $\{(\hat{z}_i,z_i)\}_{i=0}^n$ are statistically exchangeable.
\textb{As in \secref{background}, the calibration data must be distinct from the data used to train $h(\cdot)$ or any other other model.}


 
Our general approach is as follows.
Using $\{(x_i,y_i)\}_{i=1}^n$ , we compute the image recovery $\hat{x}_i=h(y_i)$ and the corresponding true FRIQ $z_i=m(\hat{x}_i,x_i)$ for each $i=1,\dots,n$.
Then we construct an estimator $f(\cdot)$ that produces an FRIQ estimate $\hat{z}_i=f(u_i)$ for some choice of $u_i$.
Several choices of $f(\cdot)$ and $u_i$ will be described in the sequel.
We then collect the results into the set $d\cal=\{(\hat{z}_i,z_i)\}_{i=1}^n$ and calibrate the $\lambda$ parameter of the FRIQ prediction interval $\mc{C}_\lambda(\hat{z}_i)$ using CP.

We now describe our choice of prediction interval $\mc{C}_\lambda(\cdot)$.
In the sequel, we will refer to those metrics $m(\cdot,\cdot)$ for which a higher value indicates better image quality (e.g., PSNR, SSIM) as higher-preferred (HP) metrics, and those for which a lower value indicates better image quality (e.g., LPIPS, DISTS) as lower-preferred (LP) metrics.
We choose to construct the prediction set for the $i$-th sample as 
\begin{align}
\mc{C}_{\lambda}(\hat{z}_i) = [\beta(\hat{z}_i,\lambda), \infty) 
\text{~~for HP metrics and~~}
\mc{C}_{\lambda}(\hat{z}_i) = (-\infty, \beta(\hat{z}_i,\lambda) ] 
\text{~~for LP metrics}
\label{eq:pred_set} ,
\end{align}
where we choose the lower/upper bound $\beta(\cdot,\cdot)$ as 
\begin{align}
\beta(\hat{z}_i,\lambda) = \hat{z}_i - \lambda
\text{~~for HP metrics and~~}
\beta(\hat{z}_i,\lambda) = \hat{z}_i + \lambda
\text{~~for LP metrics}
\label{eq:bounds} .
\end{align}
\textb{Although our prediction intervals have infinite width, they ensure that the empirical miscoverage in \eqref{empirical_miscoverage} is monotonically non-increasing in $\lambda$, as required by \citet{Angelopoulos:22}.}

By calibrating the bound parameter $\lambda$ as $\hat{\lambda}(d\cal)$ using \eqref{lambda_hat}, we obtain the following marginal coverage guarantee for the test sample $(\hat{Z}_0, Z_0)$:
\begin{equation}
\Pr \big\{ Z_{0} \in \mc{C}_{\hat{\lambda}(D\cal)}(\hat{Z}_0) \big\} \geq 1-\alpha 
\label{eq:friq_coverage} ,
\end{equation}
which holds when $\{(Z_0,\hat{Z}_0),(Z_1,\hat{Z}_1), \dots,(Z_n,\hat{Z}_n)\}$ are exchangeable \citep{Angelopoulos:22}.
In particular, $\beta(\hat{Z}_0,\hat{\lambda}(D\cal))$ lower-bounds the unknown true HP metric value $Z_0$, or upper-bounds the unknown true LP metric value $Z_0$, with probability at least $1-\alpha$, where $\alpha$ is selected by the user.
A smaller error-rate $\alpha$ will tend to yield a looser bound, but---importantly---the coverage guarantee \eqref{friq_coverage} will hold for any chosen $\alpha\in(0,1)$.
In the sequel, we will refer to $\beta(\hat{z}_0,\hat{\lambda}(d\cal))$ as the ``conformal bound'' on $z_0$.
Note that the conformal bound can ``adapt'' to the test measurements $y_0$ and reconstruction $\hat{x}_0$ through  $\hat{z}_0=f(u_0)$ for appropriate choices of $f(\cdot)$ and $u_0$.

Below we describe different ways to construct $f(\cdot)$ and $u_0$, which in turn yield conformal bounds with different properties.
\Secref{exchange} investigates violations of the exchangeability assumption.

\subsection{A non-adaptive bound on recovered-image FRIQ} \label{sec:nonadaptive}

As a simple baseline, we start with the choice $f(\cdot)=0$. 
In this case, $u_0$ is inconsequential and $\hat{z}_0 = 0$, and so the conformal bound $\beta(\hat{z}_0,\hat{\lambda}(d\cal))$ will depend on the calibration set $d\cal$ but not the test measurements $y_0$ or reconstruction $\hat{x}_0$. 
We refer to such bounds as ``non-adaptive.''
As we demonstrate in \secref{experiments}, non-adaptivity leads to conservative bounds. 
Still, this non-adaptive bound is valid in the sense of guaranteed marginal coverage \eqref{friq_coverage} under the exchangeability assumption.

\begin{figure}
    \centering
    \includegraphics[width=1\linewidth]{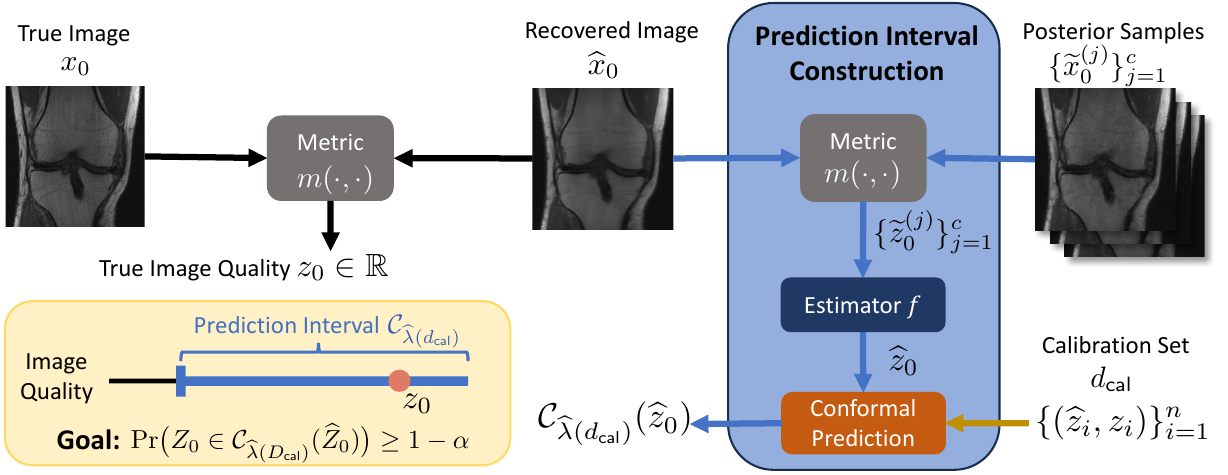}
    \caption{Overview of method: Given a recovery $\hat{x}_0$ of true image $x_0$, approximate posterior samples $\{ \tilde{x}_0\of{j} \}_{j=1}^{c}$, and a calibration set $d\cal$, we construct a prediction interval $\mc{C}_{\hat{\lambda}(d\cal)}\big(\hat{z}_0 \big)$ that is guaranteed to contain the unknown true FRIQ $z_0=m(\hat{x}_0,x_0)$ with probability at least $1-\alpha$.}
    \label{fig:overview}
\end{figure}

\subsection{Intuitions on constructing adaptive FRIQ bounds} \label{sec:intuition}

Our approach to constructing adaptive FRIQ bounds is based on the following probabilistic viewpoint. 
Conditioned on the observed measurements $y_0$, we can model the unknown FRIQ as $Z_0 = m(\hat{x}_0, X_0)$ for $\hat{x}_0=h(y_0)$ and $X_0 \sim p_{X_0|Y_0}(\cdot| y_0)$.
The distribution $p_{X_0|Y_0}(\cdot| y_0)$ is often referred to as the posterior distribution on $X_0$ given the measurements $Y_0=y_0$. 

Let us first consider the ideal and unrealistic case that the $y_0$-conditional FRIQ distribution $p_{Z_0|Y_0}(\cdot| y_0)$ is known. 
And let's consider the case of HP metrics, noting that all of our arguments can be easily modified to cover LP metrics.
If $p_{Z_0|Y_0}(\cdot| y_0)$ was known, then constructing a lower-bound $\beta$ on $Z_0$ that holds with probability $\geq 1-\alpha$ could be directly accomplished by finding the $\beta\in\Real$ that satisfies $\Pr\{Z_0\geq \beta|Y_0\!=\!y_0\}\geq 1-\alpha$, which is known as 
the $\alpha$th quantile of $Z_0|Y_0\!=\!y_0$.

Now suppose that the distribution of $Z_0|Y_0\!=\!y_0$ was unknown, but instead one had access to an infinite number of perfect posterior image samples $\{ \tilde{x}_0\of{j} \}_{j=1}^{\infty}$. 
By ``perfect'' we mean that $\tilde{x}_0\of{j}$ are independent realizations of $X_0|Y_0\!=\!y_0$.
From them, one could construct posterior FRIQs $\{ \tilde{z}_0\of{j} \}_{j=1}^{c}$ using $\tilde{z}_0\of{j} \defn m(\hat{x}_0, \tilde{x}_0\of{j})$.
Importantly, $\{z_0, \tilde{z}_0\of{1}, \tilde{z}_0\of{2}, \tilde{z}_0\of{3}, \dots\}$ are i.i.d.\ realizations of $Z_0|Y_0\!=\!y_0$.
Thus, to construct a lower bound $\beta$ on $Z_0|Y_0\!=\!y_0$ that holds with probability $1-\alpha$, one could use the empirical quantile of $\{\tilde{z}_0\of{j}\}$ , i.e.,
\begin{align}
\beta= \lim_{c\rightarrow\infty} \EmpQuant\big(\alpha, \{\tilde{z}_0\of{j}\}_{j=1}^c\big) 
\label{eq:empquant},
\end{align}
which converges to the $\alpha$th quantile of $Z_0|Y_0\!=\!y_0$
\citep{Fristedt:Book:13}.

In practice, one will not have access to an infinite number of perfect posterior image samples.
However, it is not difficult to obtain a \emph{finite} number of \emph{approximate} posterior samples $\{\tilde{x}_0\of{j}\}_{j=1}^c$. 
From them, one could 
estimate the $\alpha$th quantile of $Z_0|Y_0\!=\!y_0$
and subsequently calibrate that (imperfect) estimate using conformal prediction.
Two such strategies are described below.

\subsection{An adaptive bound on recovered-image FRIQ} \label{sec:adaptive}

Suppose that, for each $i\in\{0,1,\dots,n\}$, we have access to $c\geq 1$ approximate posterior image samples $\{\tilde{x}_i\of{j}\}_{j=1}^c$ produced by a black-box posterior image sampler such as those listed in \secref{introduction}.
Guided by the intuitions from \secref{intuition}, we propose the following for HP metrics.
For each $i$, we first compute the corresponding approximate posterior FRIQs $\{\tilde{z}_i\of{j}\}_{j=1}^c$ using $\tilde{z}_{i}\of{j} = m(\hat{x}_{i}, \tilde{x}_{i}\of{j})$ and then set $\hat{z}_i$ at their empirical quantile
\begin{align}
    \hat{z}_{i}  = \EmpQuant\big(\alpha, \{\tilde{z}_{i}\of{j}\}_{j=1}^c\big)
    =f(u_i)
    \text{~~for~~}
    \begin{cases}
    f(\cdot)=\EmpQuant(\alpha,\cdot)\\
    u_i = [\tilde{z}_{i}\of{1},\dots,\tilde{z}_i\of{c}]\tran \in\Real^c.
    \end{cases}
    \label{eq:conformal_empquant}
\end{align}
We then use $d\cal = \{(\hat{z}_i, z_i)\}_{i=1}^{n}$ to calibrate the bound parameter $\lambda$ using \eqref{lambda_hat}, yielding $\hat{\lambda}(d\cal)$.
Finally, we plug this $\lambda$ and $\hat{z}_0$ into \eqref{bounds} to get $\beta(\hat{z}_0,\hat{\lambda}(d\cal))$, which is our conformal bound on the true FRIQ $z_0$.
From \secref{background}, we know that this conformal bound satisfies the coverage guarantee \eqref{friq_coverage} under the exchangeability assumption.
Furthermore, it adapts to the measurements $y_0$ and reconstruction $\hat{x}_0$ through their effect on $\hat{z}_0$ and $\{\widetilde{z}_0^{(j)}\}_{j=1}^c$, unlike the non-adaptive bound from \secref{nonadaptive}.
We refer to these conformal bounds as the ``quantile'' bounds.

Recalling \secref{intuition}, one could interpret $\hat{z}_0$ as a rough estimate of the $\alpha$th quantile of $Z_0|Y_0=y_0$ and $\hat{\lambda}(d\cal)$ as an additive correction that accounts for the finite and approximate nature of the posterior image samples $\{\tilde{x}_0\of{j}\}_{j=1}^c$ used to construct $\hat{z}_0$.
For LP metrics, we would instead compute the $(1-\alpha)$-empirical quantile in \eqref{conformal_empquant}.
\Figref{overview} illustrates the overall methodology.

\subsection{A learned adaptive bound on recovered-image FRIQ} \label{sec:improved}

In \secref{intuition}, we reasoned that the $\alpha$th quantile of $Z_0|Y_0\!=\!y_0$ yields a valid HP FRIQ bound, but we noted that this quantile is not directly observable.
Thus, in \secref{adaptive}, we used the $\alpha$th empirical quantile of $\{\tilde{z}_i\of{j}\}_{j=1}^c$ as a rough estimate ``$\hat{z}_i$'' of the desired quantile, after which we used CP to correct this estimate and obtain a valid HP FRIQ bound.
However, it is well known from the CP literature that inaccurate base estimators cause loose conformal bounds \citep{Angelopoulos:FTML:23}.
Thus, in this section, we aim to improve our estimate of the $\alpha$th quantile of $Z_0|Y_0\!=\!y_0$.%

Inspired by conformalized quantile regression \citep{Romano:NIPS:19}, we propose to estimate the $\alpha$th quantile of $Z_0|Y_0=y_0$ using
\begin{align}
\hat{z}_i = f(u_i;\theta)
~~\text{with}~~ u_i= [\tilde{z}_i\of{1},\dots,\tilde{z}_i\of{c}]\tran \in \Real^c
\label{eq:predictor} ,
\end{align}
where $\theta$ are predictor parameters trained using quantile regression (QR) \citep{Koenker:ECON:78}.
An example $f(\cdot;\theta)$ is given in \appref{training_details}.
In the case of an HP metric, this manifests as
\begin{align}
\arg\min_{\theta} \sum_{i=n+1}^{n+n\train} \big( \alpha \max(0,z_i - \hat{z}_i(\theta)) + (1-\alpha) \max(0,\hat{z}_i(\theta) - z_i)  \big) 
+ \gamma \rho(\theta)
\label{eq:QR} ,
\end{align}
using a training set $d\train = \{ ( u_i, z_i) \}_{i=n+1}^{n+n\train}$ that is independent of the calibration samples $\{ (u_i, z_i) \}_{i=1}^n$ and test sample $(u_{0}, z_{0})$. 

The first term in \eqref{QR} is the pinball loss \citep{Koenker:ECON:78}, which encourages an $\alpha$-fraction of training samples to violate the HP bound $\hat{z}_i\leq z_i$.
The $\rho(\cdot)$ term in \eqref{QR} is regularization that avoids overfitting $\theta$ to the training set.
The regularization weight $\gamma$ can be tuned using k-fold cross-validation.
The $\theta$-dependence of $\hat{z}_i$ is made explicit in \eqref{QR}.

Once the predictor $f(\cdot;\theta)$ is trained, it can be used to obtain the quantile estimates $\{\hat{z}_i\}_{i=0}^n$.
Then $d\cal\defn\{(\hat{z}_i,z_i)\}_{i=1}^n$ can be used to calibrate the bound parameter $\lambda$ using \eqref{lambda_hat}.
As before, the resulting conformal bound $\beta(\hat{z}_0,\hat{\lambda}(d\cal))$ will enjoy the coverage guarantee \eqref{friq_coverage} under the exchangeability assumption.
To handle LP metrics, we would swap $\alpha$ with $1-\alpha$ in \eqref{QR}.
Note that any estimation function $f(\cdot;\theta)$ can be used in \eqref{predictor} and the best choice will vary with the application.
Through the remainder of the paper, we describe these bounds as the ``regression'' bounds.

\section{Numerical experiments} \label{sec:experiments}

\begin{figure}[t]
    \centering
    \includegraphics[width=1\linewidth]{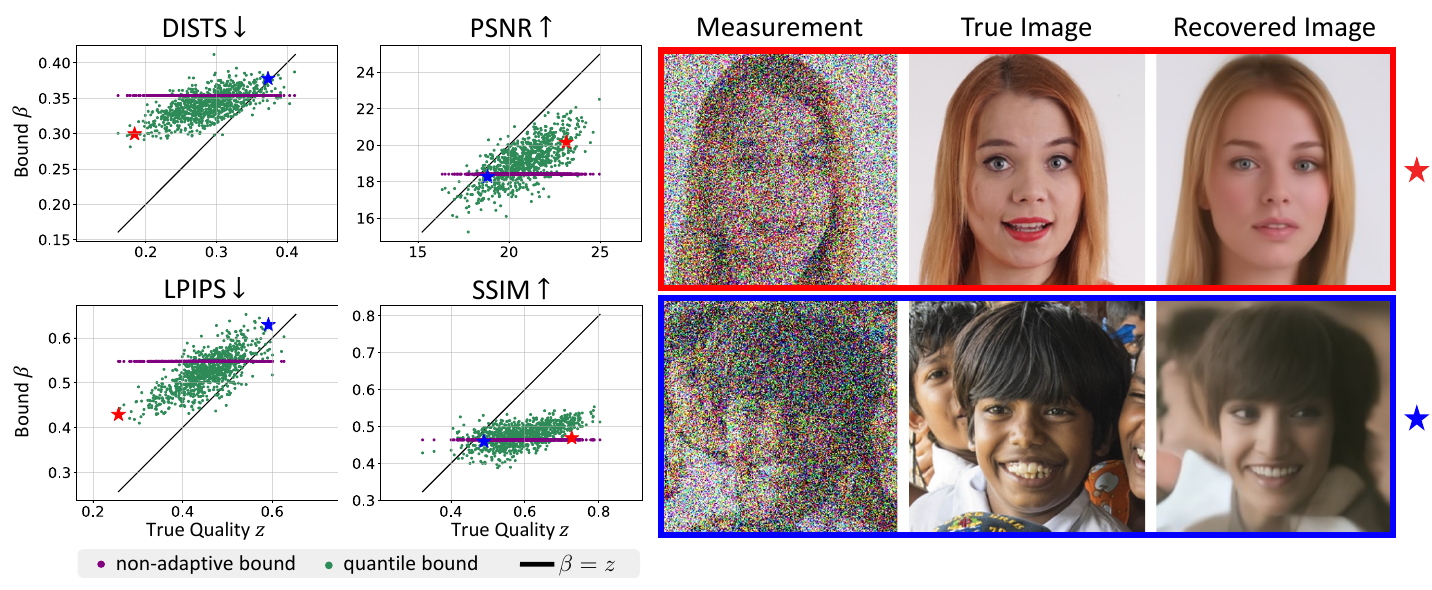}
    \caption{Scatter plots show the non-adaptive (purple) and quantile (green) bounds $\beta(\hat{z}_k, \hat{\lambda}(d\cal[t]))$ versus the true FRIQ $z_k$ over FFHQ test samples $k$. The black line shows where $\beta=z$, and a fraction $\alpha=0.05$ of samples are on the side of the line that violates the bound. The quantile bound tracks the true $z_k$ much better than the non-adaptive bound. The red and blue stars correspond to the images in the red and blue boxes: the red recovery represents better FRIQs and blue represents worse.}
    \label{fig:ddrm_qualitative}
\end{figure}

\begin{figure}
    \centering
    \includegraphics[width=1\linewidth]{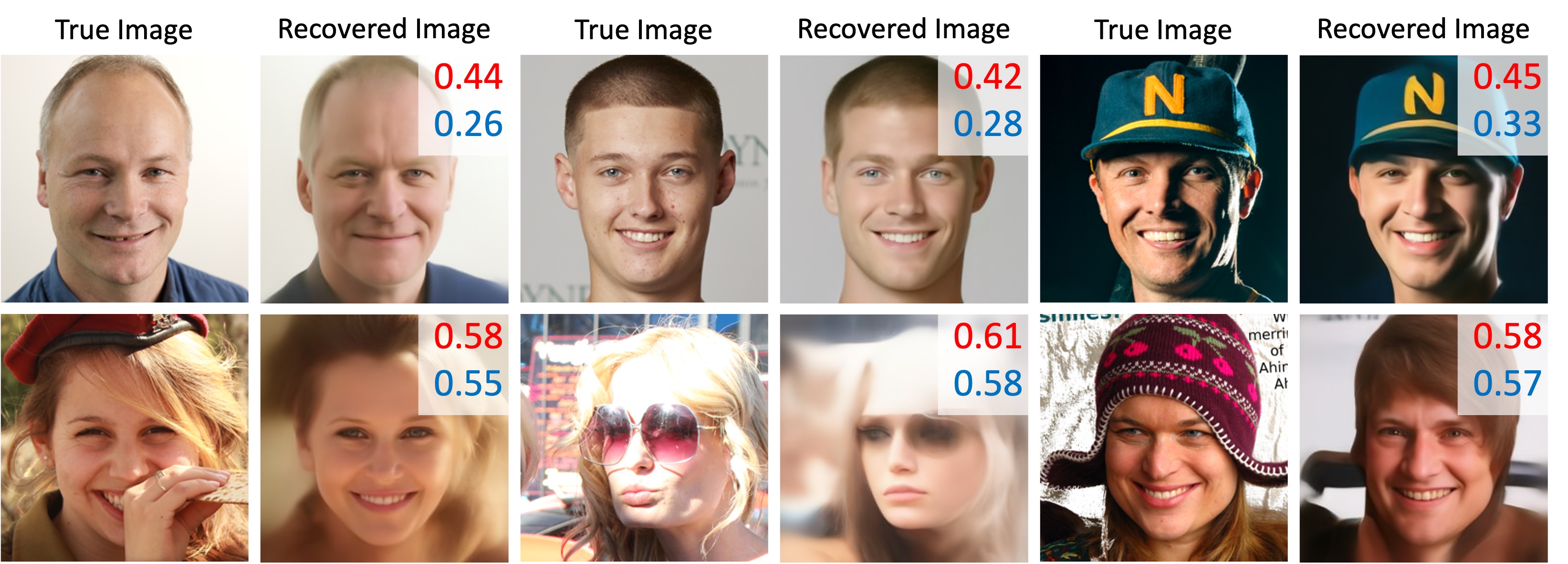}
    \caption{Examples from the FFHQ denoising experiment. 
    Top row: true image and low-LPIPS recovery. 
    Bottom row: true image and high-LPIPS recovery.
    True LPIPS reported in blue and quantile upper-bound in red.
    (Recall that LPIPS assigns lower values to better recoveries.)
    }
    \label{fig:ffhq_qualitative}
\end{figure}

\begin{figure}[t]
    \centering
    \includegraphics[width=1\linewidth]{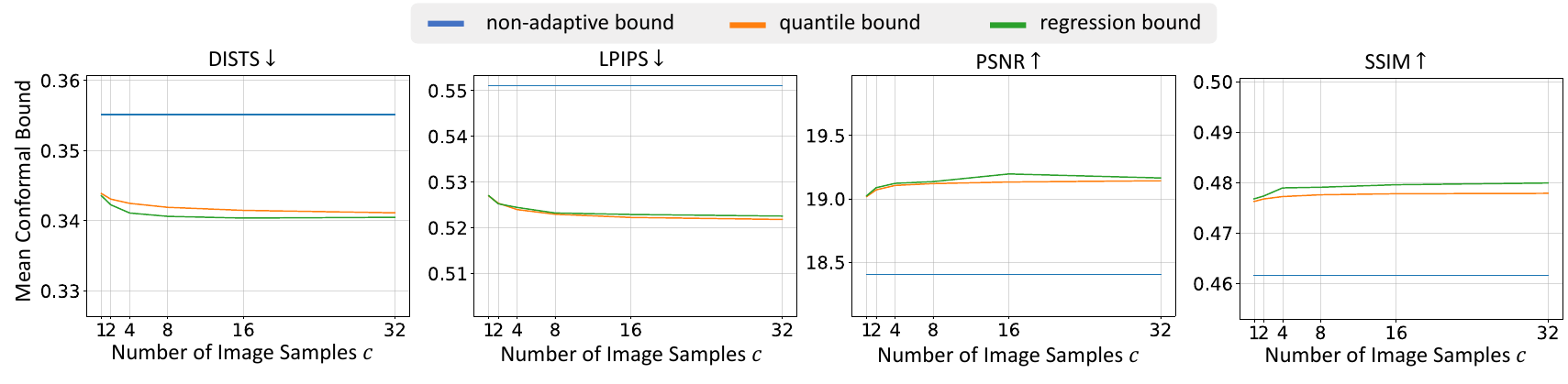}
    \caption{Mean conformal bound versus number of posterior samples $c$ for FFHQ denoising.} 
    \label{fig:ddrm_bound}
\end{figure}

\begin{figure}[t]
    \centering
    \includegraphics[width=1\linewidth]{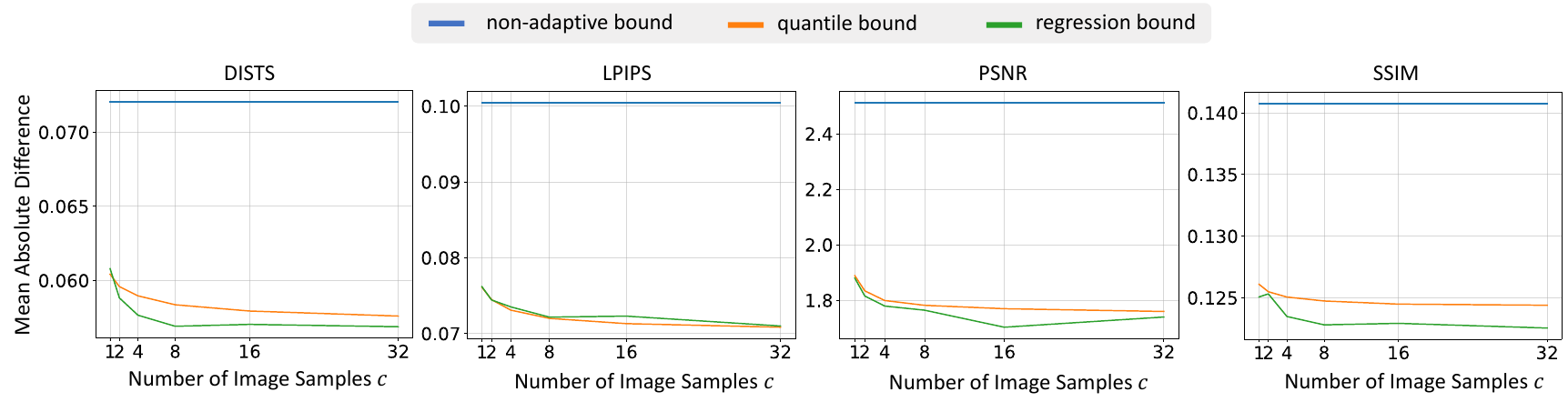}
    \caption{\textb{Mean absolute difference between the bound and true FRIQ versus number of posterior samples $c$ for FFHQ denoising.}}
    \label{fig:ddrm_abs_diff}
\end{figure}

We now consider two imaging inverse problems: image denoising and accelerated MRI.
For each, we evaluate the proposed bounds using the PSNR, SSIM \citep{Wang:TIP:04}, LPIPS \citep{Zhang:CVPR:18}, and DISTS \citep{Ding:TPAMI:20} metrics. 

\subsection{Denoising} \label{sec:denoising}
\textbf{Data:}~ For true images, we use a random subset of 4000 images from the Flickr Faces HQ (FFHQ) \citep{Karras:CVPR:19} validation dataset, to which we added white Gaussian noise of standard deviation $\sigma= 0.75$ to create the measurements $y_0$. 
The first 1000 images were used to train the predictor $f(\cdot;\theta)$ in \eqref{predictor} and the remaining 3000 were used for calibration and testing.

\textbf{Recovery:}~ To recover $\hat{x}_0$ from $y_0$, a denoising task, we use the Denoising Diffusion Restoration Model (DDRM) \citep{Kawar:NIPS:22}. 
Following \citet{Kawar:NIPS:22}, we run DDRM with a Denoising Diffusion Probabilistic Model (DDPM) \citep{Ho:NIPS:20} pretrained on the CelebA-HQ dataset \citep{Karras:ICLR:18}. 
To increase sampling diversity, we used $\eta=1$ and $\eta_b=0.5$ but set all other hyperparameters at their default values.
For each measurement $y_i$, we use one DDRM sample for the image estimate $\hat{x}_i$ and $c$ independent samples for $\{ \tilde{x}_i\of{j} \}_{j=1}^{c}$. 

\textbf{Conformal bounds:}~ We evaluate the proposed bounding methods from Secs.~\ref{sec:nonadaptive}, \ref{sec:adaptive}, and \ref{sec:improved}, which we refer to as the \textbf{non-adaptive}, \textbf{quantile}, and \textbf{regression} bounds, respectively. 
For the regression bound, we use a quantile predictor $f(\cdot,\theta)$ that takes the form of a linear spline with two knots (see \Appref{training_details} for more details).

\textbf{Validation procedure:}~ 
Because the coverage guarantee \eqref{friq_coverage} involves random calibration data and test data, we
evaluate our methods using $T$ Monte-Carlo trials. 
For each trial $t\in\{1,\dots,T\}$, we randomly select 70\% of the 3000 non-training samples to create the calibration set $d\cal[t]$ with indices $i \in \mc{I}\cal[t]$,
and we use the remaining 30\% of the non-training samples for a test fold with indices $k \in \mc{I}\test[t]$.
In particular, we compute $\hat{\lambda}$ using $d\cal[t]$ and then, for each test sample index $k$, we compute the bound $\beta(\hat{z}_k, \hat{\lambda}(d\cal[t]))$.
Finally, we average performance across the test indices of trial $t$ and then average those results across the $T$ trials.
Unless specified otherwise, we used error rate $\alpha=0.05$, $T=10\,000$, and $c=32$ samples for the adaptive bounds.

\textbf{Bound versus true metric:}~
Ideally, an FRIQ bound should track the true FRIQ in the sense that the bound is small when the true FRIQ is small and large when the true FRIQ is large.
To assess this tracking behavior,
\figref{ddrm_qualitative} shows scatter plots of the non-adaptive and quantile bounds $\beta(\hat{z}_k, \hat{\lambda}(d\cal[t]))$ versus the true FRIQ $z_k$ for the test indices $k \in \mc{I}\test[t]$ of a single Monte Carlo trial, along with the true image $x_k$ and recovery $\hat{x}_k$ for two test samples.
The sample highlighted in red has better subjective visual quality compared to the one in blue, and this ranking is reflected in both the true FRIQ metrics $z_k$ and the corresponding quantile bounds, but not the non-adaptive bound. 
\Figref{ffhq_qualitative} shows six additional samples from the FFHQ denoising experiment, three with low (true) LPIPS and three with high (true) LPIPS, along with the respective true images. 
The quantile upper-bound on LPIPS is superimposed on each recovery. 
We see that the bounds are valid in the sense that they did not under-predict the true LPIPS and adaptive in the sense that the bounding value is lower when the true LPIPS is lower. 

\textbf{Empirical Coverage:}~ 
To verify the coverage guarantee in \eqref{friq_coverage} is satisfied, we compute the empirical coverage 
\begin{align}
    \EC[t] 
    \defn \frac{1}{|\mc{I}\test[t]|} \sum_{k \in \mc{I}\test[t]} \mathds{1}\{z_k \in  \mc{C}_{\hat{\lambda}(D\cal)}(\hat{z}_k) \} 
    \label{eq:ECt},
\end{align}
for each Monte Carlo trial $t$.
In Table \ref{tab:ffhq_converages}, we report the average empirical coverage and standard error across $T=10\,000$ trials for all three methods on the FFHQ data using a target error rate of $\alpha=0.05$.
For all methods, the average empirical coverage is very close to the theoretical coverage of $1-\alpha=0.95$ regardless of the metric, demonstrating close adherence to the theory.
In \Appref{empirical_coverage}, we further demonstrate that this adherence holds independent of the choice of $c$.

\begin{table}[t]
    \centering
    \caption{Mean empirical coverage for all bounds with $\alpha=0.05$ and $T=10\,000$ on the FFHQ denoising task ($\pm$ standard error). Quantile and regression bounds are computed with $c=32$.}
    \label{tab:ffhq_converages}
    \begin{tabular}{ccccc}
        \toprule
        Bound & DISTS & LPIPS & PSNR & SSIM \\
        \midrule
        Non-adaptive  & $0.95000 \pm 0.00009$ & $0.95016 \pm 0.00009$ & $0.95004 \pm 0.00009$ & $0.95010 \pm 0.00009$ \\
        Quantile & $0.95002 \pm 0.00009$ & $0.95013 \pm 0.00009$ & $0.95003 \pm 0.00009$ & $0.95006 \pm 0.00009$ \\
        Regression & $0.95013 \pm 0.00009$ & $0.95026 \pm 0.00009$ & $0.95001 \pm 0.00009$ & $0.95006 \pm 0.00009$ \\
        \bottomrule
    \end{tabular}
\end{table}

\textbf{\textb{Bound tightness} versus bounding method and number of posterior samples $c$:}~ 
To assess the tightness of the conformal bounds, we average the bound $\beta(\hat{z}_k, \hat{\lambda}(d\cal[t]))$ over the test indices $k \in \mc{I}\test[t]$ and the Monte Carlo trials $t$ to yield the ``mean conformal bound'' (MCB).
\textb{We also average the absolute difference between the bound and true FRIQ $|z_k - \beta(\hat{z}_k, \hat{\lambda}(d\cal[t]))|$ over the test indices and Monte Carlo trials to find the ``mean absolute difference'' (MAD).}
\Figref{ddrm_bound} plots the MCB versus the number of posterior samples $c$ used for the adaptive bounds.
The figure shows that the non-adaptive bound is looser (i.e., smaller for the HP metrics PSNR and SSIM and larger for the LP metrics DISTS and LPIPS) than the two adaptive bounds.
\textb{\Figref{ddrm_abs_diff} plots the MAD versus the number of posterior samples $c$.
Similarly, this figure shows the non-adaptive bound is further from the true FRIQ compared to the quantile and regression bounds.}
For both adaptive bounds, Figs.\ \ref{fig:ddrm_bound} \textb{and \ref{fig:ddrm_abs_diff} show} only minor bound improvement with increasing $c$, suggesting that the adaptive bounds are robust to the choice of $c$, and that small values of $c$ could suffice if sample-generation was computationally expensive.
(We discuss computation time below.)

Interestingly, Figs.\ \ref{fig:ddrm_bound} \textb{and \ref{fig:ddrm_abs_diff} show} relatively little improvement when going from the quantile bound to the regression bound.
This may be due to our choice of a linear spline with two knots for $f(\cdot;\theta)$, but experiments with higher spline orders  and/or more knots did not yield improved results, and neither did experiments with XGBoost \citep{Chen:ICKDDM:16} models for $f(\cdot;\theta)$.
Additional experiments that hold the number of test samples at 900 and vary $n\train$ and $n\cal$ such that $n\train+n\cal=3100$ (see \Appref{additiona_ffhq}) also show little change in the performance of the quantile and regression bounds.
Thus, for our experimental data, the effort to train the estimation function $f(\cdot;\theta)$ from \eqref{predictor} may not be justified, given the good performance of the simple empirical-quantile estimation function $f(\cdot)$ from \eqref{conformal_empquant}.  
But the behavior may be different with other datasets.



\textbf{Computation time:}~
Using a single NVIDIA V100 GPU with 32GB of memory, 
computing a single DDRM sample takes approximately 2.73 seconds. 
Once the calibration constant $\hat{\lambda}(d\cal)$ is known, computing $c=32$ FRIQ samples $\{\tilde{z}_0\of{j}\}_{j=1}^c$ and $\beta(\hat{z}_0, \hat{\lambda}(d\cal))$ takes around 217ms, 320ms, 5ms, and 6ms for DISTS, LPIPS, PSNR, and SSIM, respectively.

\subsection{Accelerated MRI} \label{sec:MRI}
We now simulate our methods on accelerated multicoil MRI \citep{Knoll:SPM:20,Hammernik:SPM:23}. 
MRI is a medical imaging technique known for excellent soft tissue contrast without subjecting the patient to harmful ionizing radiation. 
MRI has slow scan times, though, which reduce patient throughput and comfort. 
In accelerated MRI, one collects only $1/R$ of the measurements specified by the Nyquist sampling theorem, thus speeding up the acquisition process by rate $R$.
For $R>1$, however, the inverse problem may become ill-posed, in which case one may be interested in bounding the FRIQ of the recovered image.

\textbf{Data:}~ 
We utilize the non-fat-suppressed subset of the multicoil fastMRI knee dataset \citep{Zbontar:18}, yielding 17286 training images and 2188 validation images. 
To simulate the imaging process, we retrospectively sub-sample in the spatial Fourier domain (the ``k-space'') using random Cartesian masks that give acceleration rates $R \in \{ 16,8,4,2\}$.
See \appref{mri_masks} for additional details.

\textbf{Recovery:}~ 
To generate approximate posterior samples for the adaptive bounds, we utilize the conditional normalizing flow (CNF) from \citet{Wen:ICML:23}.
Although in \appref{cnf} we investigate what happens when the same CNF is also used as the recovery network $h(\cdot)$, in this section we use the well-known E2E-VarNet \citep{Sriram:MICCAI:20}---a deterministic reconstruction approach, for $h(\cdot)$.

Both networks are trained to work well with all four acceleration rates $R$. 
(See \appref{training_details} for training details.)
Similar to \secref{denoising}, we found that the regression bound did not provide significant gain over the quantile bound and so, to streamline our discussion, we consider only the quantile and non-adaptive bounds for MRI.
As before, we evaluate performance over $T=10\,000$ Monte Carlo trials with a random 70\% calibration and 30\% test split of the validation data.
All experiments use an error-rate $\alpha=0.05$.
Methods are separately calibrated for each acceleration rate.

\begin{figure}[t]
    \centering
    \includegraphics[width=1\linewidth]{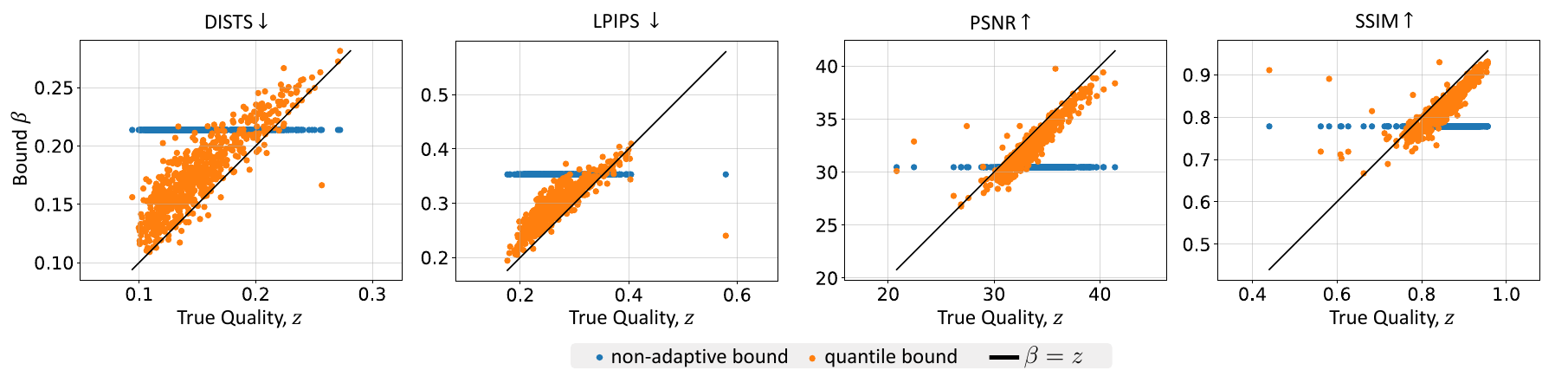}
    \caption{Scatter plots show the non-adaptive (blue) and quantile (orange) bounds $\beta(\hat{z}_k, \hat{\lambda}(d\cal[t]))$ versus the true FRIQ $z_k$ over MRI test indices $k \in \mc{I}\test[t]$ at acceleration $R=8$. The black line shows where $\beta=z$. A fraction of $\alpha=0.05$ samples are on the side of the line that violates the bound. Note that the quantile bound tracks the true $z_k$ much better than the non-adaptive bound.}
    \label{fig:mri_scatter}
\end{figure}

\textbf{Bound versus true-metric:}~
\Figref{mri_scatter} shows scatter plots of the true FRIQ $z_k$ versus the non-adaptive and quantile bounds $\beta(\hat{z}_k, \hat{\lambda}(d\cal[t]))$ for the test indices $k \in \mc{I}\test[t]$ in a single Monte-Carlo trial $t$.
The results are shown for $R=8$ acceleration and $c=32$ samples in the adaptive bounds.
Except for a few outliers, the quantile bound closely tracks the true FRIQ $z_k$, demonstrating good adaptivity, while the non-adaptive bounds remain constant with $z_k$. 

\begin{figure}[t]
    \centering
    \begin{minipage}{0.45\linewidth}
        \centering
        \includegraphics[width=1.0\linewidth,trim= 0 10 0 0,clip]{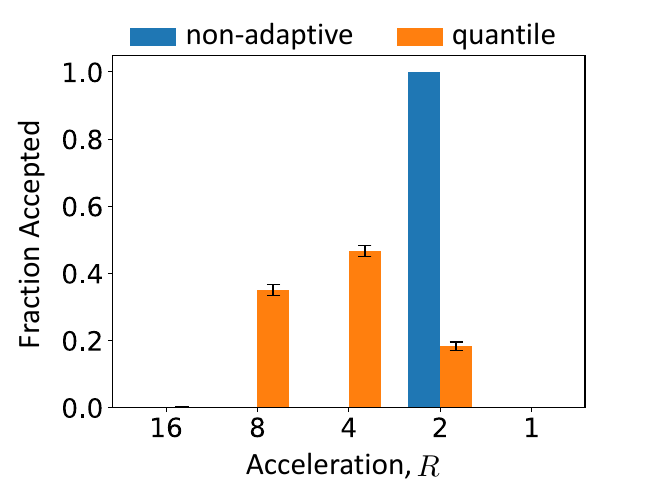}
        \caption{Fraction of accepted slices versus final acceleration rate for multi-round MRI using DISTS with $\tau=0.16$. Error bars show standard deviation.
        }
        \label{fig:slices_remaining_e2ecnf}
    \end{minipage}
    \hfill
    \begin{minipage}{0.45\linewidth}
        \centering
        \includegraphics[width=1.0\linewidth,trim= 0 10 0 0,clip]{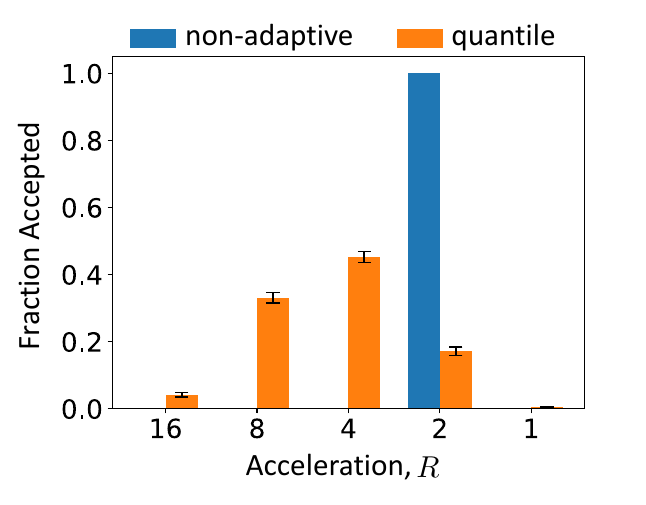}
        \caption{Fraction of accepted slices versus final acceleration rate for multi-round MRI using PSNR with $\tau=33$dB. Error bars show standard deviation.
        }
        \label{fig:slices_remaining_e2ecnf_psnr}
    \end{minipage}
    
\end{figure}

\begin{table}[t]
    \centering
    \captionof{table}{Average results for a multi-round MRI simulation where measurement collection stops once the DIST bound is below a user-set threshold $\tau$. Results shown for $T=10\,000$ trials using $\alpha=0.05$, $\tau=0.16$, and $c=32$ ($\pm$ standard error).
    }
    \resizebox{\columnwidth}{!}{%
    \begin{tabular}{|c|c|c|c|c|c|c|}
        \hline
        Method 
        & \multicolumn{1}{|p{2cm}|}{\centering Average \\ Acceleration}  
        & \multicolumn{1}{|p{2cm}|}{\centering Acceptance \\ Empirical \\ Coverage}
        & \multicolumn{1}{|p{2cm}|}{\centering $R=16$ \\ Empirical \\ Coverage}
        & \multicolumn{1}{|p{2cm}|}{\centering $R=8$ \\ Empirical \\ Coverage}
        & \multicolumn{1}{|p{2cm}|}{\centering $R=4$ \\ Empirical \\ Coverage}
        & \multicolumn{1}{|p{2cm}|}{\centering $R=2$ \\ Empirical \\ Coverage}
        \\
        \hline
        Non-adaptive & $2.000 \pm 0.000$ & $0.9504 \pm 0.0001$ & $0.9505 \pm 0.0001$ & $0.9505 \pm 0.0001$ & $0.9505 \pm 0.0001$ & $0.9506 \pm 0.0001$\\
        Quantile & $3.973 \pm 0.001$ & $0.9323 \pm 0.0001$ & $0.9506 \pm 0.0001$ & $0.9505 \pm 0.0001$ & $0.9505 \pm 0.0001$ & $0.9503 \pm 0.0001$ \\
        \hline
    \end{tabular}
    }
    \label{tab:multi_round_e2ecnf}
\end{table}

\begin{table}[t]
    \centering
    \captionof{table}{Average results for a multi-round MRI simulation where measurement collection stop once the PSNR bound is above a user-set threshold $\tau$. Results shown for $T=10\,000$ trials with $\alpha=0.05$, $\tau=33.0$ dB, and $c=32$ ($\pm$ standard error).
    }
    \resizebox{\columnwidth}{!}{%
    \begin{tabular}{|c|c|c|c|c|c|c|}
        \hline
        Method 
        & \multicolumn{1}{|p{2cm}|}{\centering Average \\ Acceleration}  
        & \multicolumn{1}{|p{2cm}|}{\centering Acceptance \\ Empirical \\ Coverage}
        & \multicolumn{1}{|p{2cm}|}{\centering $R=16$ \\ Empirical \\ Coverage}
        & \multicolumn{1}{|p{2cm}|}{\centering $R=8$ \\ Empirical \\ Coverage}
        & \multicolumn{1}{|p{2cm}|}{\centering $R=4$ \\ Empirical \\ Coverage}
        & \multicolumn{1}{|p{2cm}|}{\centering $R=2$ \\ Empirical \\ Coverage}
        \\
        \hline
        Non-adaptive & $2.000 \pm 0.000$ & $0.9503 \pm 0.0001$  & $0.9504 \pm 0.0001$ & $0.9504 \pm 0.0001$ & $0.9504 \pm 0.0001$ & $0.9504 \pm 0.0001$\\
        Quantile & $4.048 \pm 0.001$ & $0.9514 \pm 0.0001$  & $0.9503 \pm 0.0001$ & $0.9505 \pm 0.0001$ & $0.9504 \pm 0.0001$ & $0.9503 \pm 0.0001$\\
        \hline
    \end{tabular}
    }
    \label{tab:multi_round_e2ecnf_psnr}
\end{table}

\begin{figure}
    \centering
    \includegraphics[width=1\linewidth]{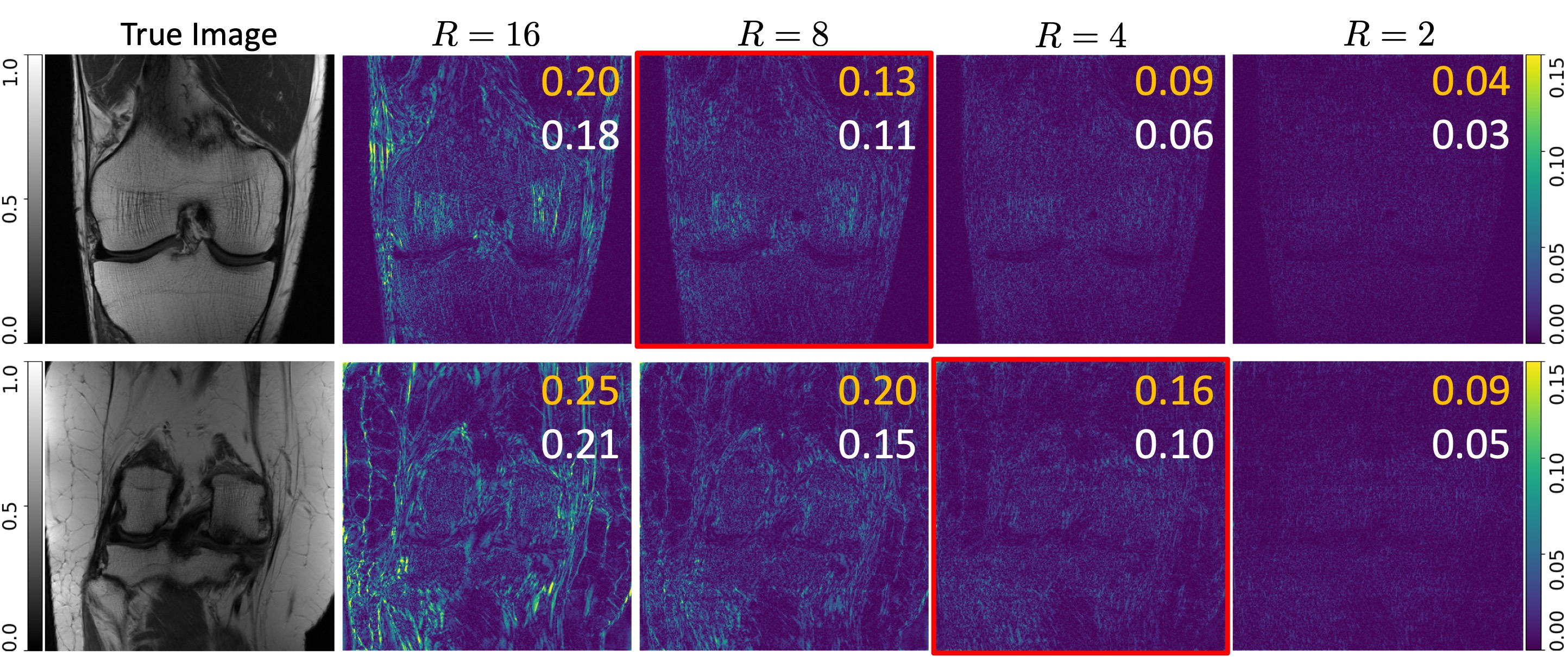}
    \caption{Examples of the multi-round MRI measurement procedure with DISTS at $\alpha=0.05$, $\tau=0.16$, and $c=32$. Error images at each acceleration $R$ are shown with the quantile bound (orange) and true metric (white). The red box indicates the measurement round at which the bound falls below the threshold $\tau$ and the measurement procedure concludes.}
    \label{fig:multi_round_qualitative}
\end{figure}

\textbf{Multi-round measurement:}
To showcase the practical impact of our bounds, we adapt the multi-round measurement protocol from \citet{Wen:24}, where measurements are collected over multiple rounds until the uncertainty bound falls below a threshold.
In our setting, measurements are first collected at acceleration $R=16$, an image recovery is computed, and a conformal upper-bound on its DISTS is computed.
If the bounding value is lower than a pre-determined threshold $\tau$, signifying that the recovery is (with probability $1-\alpha$) of sufficient diagnostic quality \citep{Kastryulin:IA:23}, then measurement collection stops.
If not, additional measurements are collected and combined with the previous ones to yield an acceleration of $R=8$, and the process repeats.
We allow up to five measurement rounds, corresponding to final accelerations of $R \in \{16,8,4,2,1\}$.

Once again, we report average results across $T=10\,000$ trials.
We set the DISTS acceptance threshold at $\tau=0.16$, which requires the non-adaptive approach to use acceleration $R=2$ in order to guarantee $1-\alpha$ empirical coverage.
\Figref{slices_remaining_e2ecnf} plots the fraction of test image slices accepted by the multi-round protocol at each acceleration rate $R$ with $\tau=0.16$.
With the quantile bound, the measurements stop after three of fewer rounds (i.e., $R\geq4$) in more than 80\% of the cases.
With the non-adaptive bound, the measurements stop after four rounds (i.e., $R=2$) in all cases.
\Tabref{multi_round_e2ecnf} shows that, with the quantile bound, the multi-round protocol attains an average acceleration of $R=3.973$, which far surpasses the $R=2$ acceleration achieved with the non-adaptive bound. 
\textb{Since a separate conformal predictor is calibrated for each acceleration rate, \Tabref{multi_round_e2ecnf} shows that each predictor satisfies the coverage guarantees \eqref{friq_coverage} when evaluated on all test samples at the corresponding acceleration rate.
We also show the acceptance empirical coverage, which computes the coverage when the slices are accepted.
Despite only having coverage guarantees for individual acceleration rates, the acceptance empirical coverage is very close to $1-\alpha$.
}
\Figref{multi_round_qualitative} shows examples of the image-error, the true DISTS, and its quantile upper-bound for each measurement round. 
With the threshold set at $\tau=0.16$, the example on the top would collect two rounds of measurements (i.e., $R=8$) while the example at the bottom would collect three rounds of measurements (i.e., $R=4$), as demarcated by the red squares.
See \appref{add_mri_experiments} for additional qualitative results.

We now repeat the multi-round experiment using the (perhaps more familiar) metric of PSNR. 
We set the acceptability threshold at $\tau=33.0$ dB, which requires the non-adaptive approach to use acceleration $R=2$ to guarantee $1-\alpha$ empirical coverage.
Similar to what happened when DISTS was used, \figref{slices_remaining_e2ecnf_psnr} shows that the quantile bound allows a large proportion of the slices to be collected at $R\geq4$.
In \tabref{multi_round_e2ecnf_psnr}, we see that this results in an average accepted acceleration rate of $R=4.048$, over twice the acceleration achieved with the non-adaptive bound. 

\textbf{Computation time:}~
The E2E-VarNet takes approximately 104ms to generate a single posterior sample, while the CNF take about 1.22 seconds to generate 32 posterior samples (corresponding to $c=32$) on a single NVIDIA V100. 
The computation time of the metrics and bounds is on par with the times reported for the FFHQ experiments. 

\subsection{Investigation of distribution shift in MRI} \label{sec:exchange}

\begin{figure}
    \centering
    \includegraphics[width=1\linewidth]{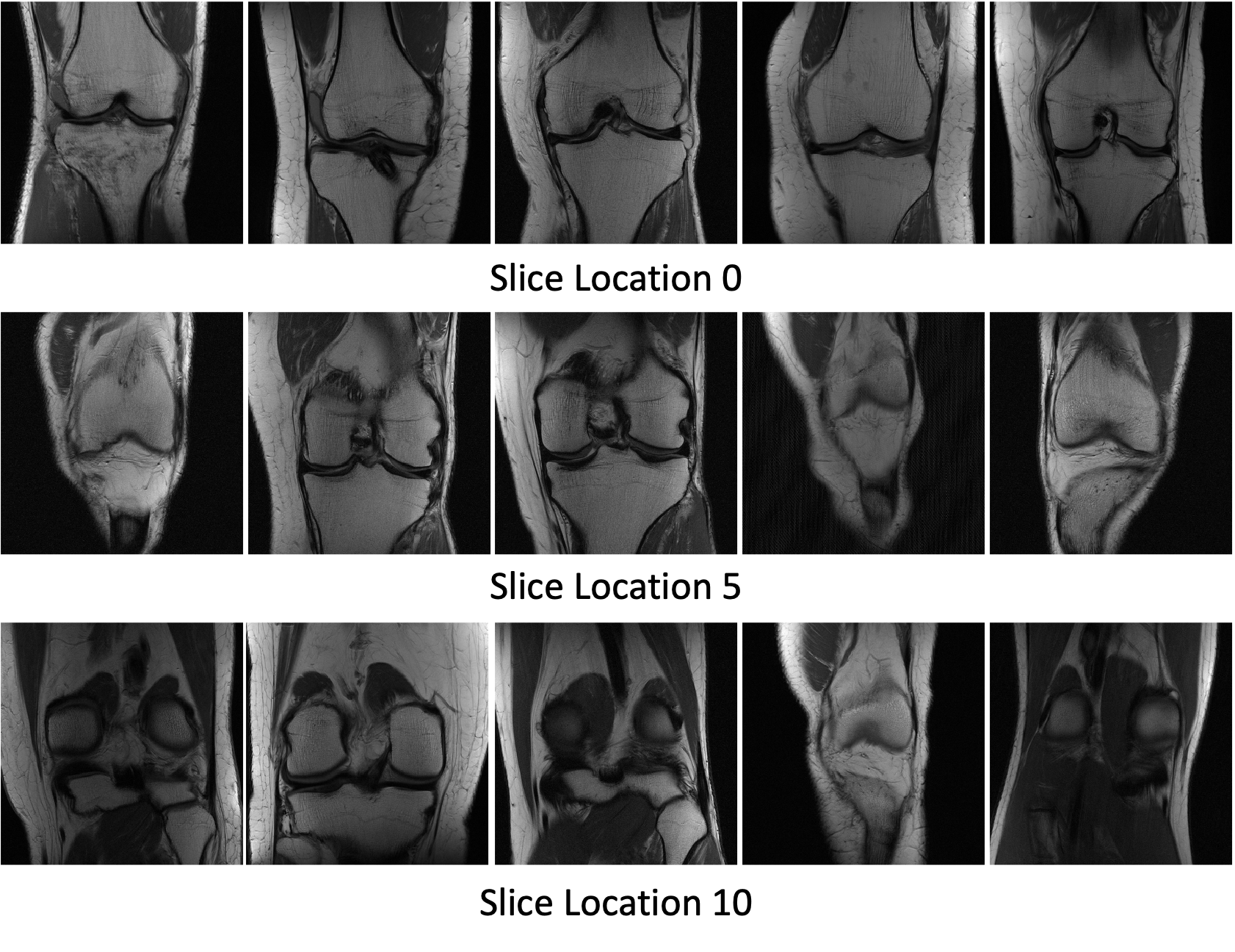}
    \caption{Qualitative examples of images from different slice locations. Slice location $0$ indicates the center slice of a volume while larger slice locations are further towards the edges of a volume.}
    \label{fig:slice_loc_qualitative}
\end{figure}

\begin{figure}[t]
    \centering
    \begin{minipage}{0.6\linewidth}
        \centering
        \includegraphics[width=1.0\linewidth,trim= 0 0 0 0,clip]{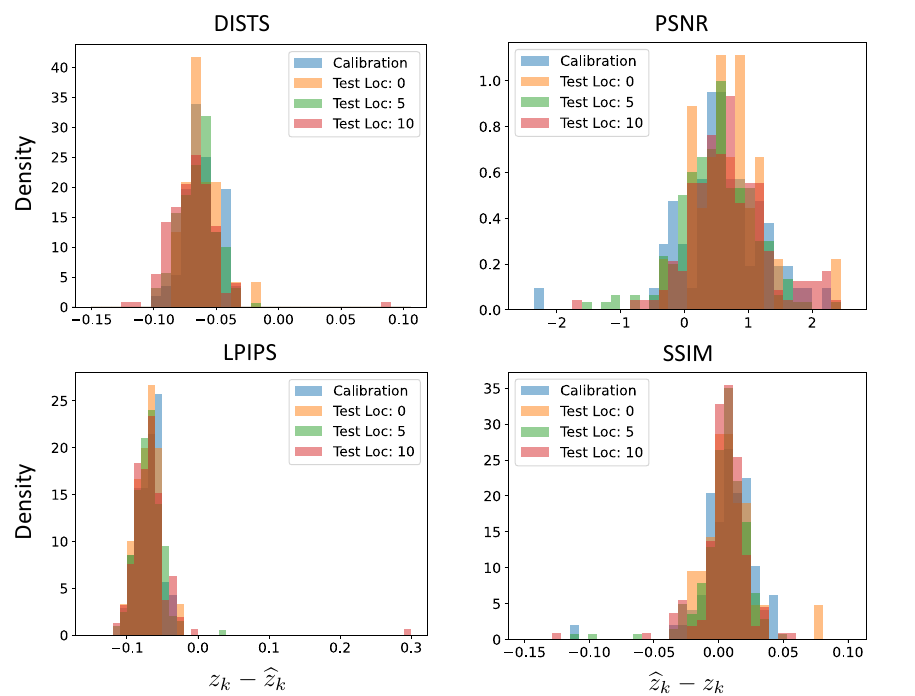}
        \caption{Histograms of the difference between the true FRIQ $z_k$ and the FRIQ estimate $\hat{z}_k$ for test indices $k$ in the test fold $\mc{I}\test[t]$ of a single trial. Histograms are shown for test slice locations $l=0,5,10$. Note the increasing shift in distribution from the calibration set as $l$ increases.}
        \label{fig:sliceloc_density_e2ecnf}
    \end{minipage}
    \hfill
    \begin{minipage}{0.38\linewidth}
        \centering
        \includegraphics[width=1.0\linewidth,trim= 0 10 0 0,clip]{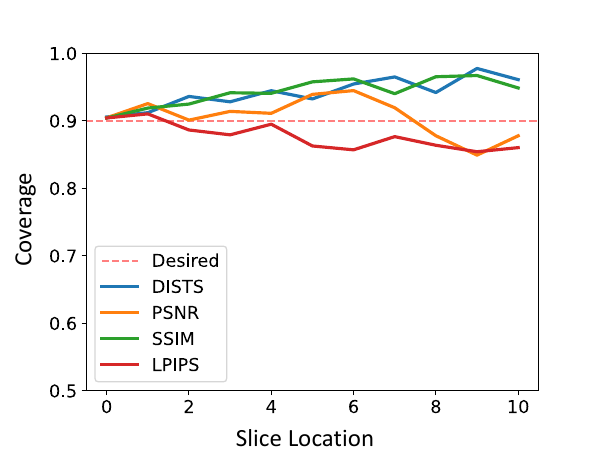}
        \caption{The average empirical coverage across $T=10000$ trials for test sets at different slice locations. All trials are calibrated with images from slice location $0$ with $\alpha=0.1$, $R=8$, and $c=32$. 
        }
        \label{fig:coverage_v_sliceloc_e2ecnf}
    \end{minipage}
\end{figure}

As previously mentioned, a general limitation of CP methods like \cite{Angelopoulos:22} is the requirement of exchangeability, which in our case applies to the pairs $\{(\hat{Z}_i,Z_i)\}_{i=0}^n$.
This requirement may be violated when there is a distributional shift between the test data $(x_0,y_0)$ and the calibration data $\{(x_i,y_i)\}_{i=1}^n$, which can then cause a distributional shift between the corresponding FRIQ quantities $(\hat{z}_0,z_0)$ and $\{(\hat{z}_i,z_i)\}_{i=1}^n$.

In the case of MRI, such distributional shifts may arise for various reasons, some of which would be easy to prevent while others would be more difficult.
For example, if the CP method was calibrated on knee images, one would not want to immediately test on brain images, but instead recalibrate a CP method on brain images.
Likewise, if the CP method was calibrated with data from one manufacturer and/or strength of scanner, then it would be best to test on data from the same manufacturer and/or strength of scanner.
Still, due to limited calibration data, situations may arise where a distribution shift is inevitable.
Thus, we perform a study to analyze the sensitivity of our proposed method to distribution shifts.

For this study, we use the validation fold of the non-fat-suppressed multicoil fastMRI knee dataset \citep{Zbontar:18}, which contains 100 3D volumes.
A volume contains all the images collected for a single patient, with each image showing a different slice of the knee (from front to back). 
To induce a realistic yet controllable distribution shift, we choose calibration images from only the center slices of these volumes, and refer to the center slices as ``location $l=0$.''
We then create one test set with images from slice locations $l=0$, another test set with images from slice location $l=1$, and so on, until slice location $l=10$ (which typically corresponds to an edge slice).
Example images from various slice locations are shown in \figref{slice_loc_qualitative}.

We first evaluate the coverage of the quantile bound using $T=10\,000$ Monte Carlo trials, error-rate $\alpha=0.1$, acceleration $R=8$, an E2E-VarNet \citep{Sriram:MICCAI:20} sample for $\hat{x}_i$, and $c=32$ posterior samples for $u_i$. 
For each trial $t\in \{1,\dots,T\}$, we construct the calibration set by randomly sampling $70$ of the $100$ center slices.
For the same $t$, we form the test data at location $l=0$ using the remaining $30$ slices, and we form the test data at locations $l>0$ by randomly sampling $30$ of the $200$ available slices. 
\Figref{coverage_v_sliceloc_e2ecnf} plots the mean empirical coverage over the $T$ trials as a function of test slice location $l$. 
As expected, the desired $1-\alpha$ coverage is met when $l=0$.
However, the behavior of the empirical coverage for $l>0$ varies depending on the metrics. 
The coverage for LPIPS tends to decrease slightly as the slice location $l$ increases, and the coverage for PSNR only falls below $1-\alpha$ after $l=7$.
Surprisingly, for the DISTS and SSIM metrics, the coverage remains well above $1-\alpha$ for all slice locations, suggesting the bounds remain valid, but become slightly over-conservative for $l>0$.
Overall, the results demonstrate our bounds are quite robust to small distributional shifts with only a minor loss in coverage for certain metrics. 

To visualize the distribution shift versus test location $l$, we consider the difference between the true FRIQ $z_k$ and the FRIQ estimate $\hat{z}_k$ for each test index $k \in \mc{I}\test[t]$ in a single trial $t$.
This difference is $z_k - \hat{z}_k$ for LP metrics and $\hat{z}_k - z_k$ for HP metrics.
\Figref{sliceloc_density_e2ecnf} shows the histogram of this difference for test locations $l\in\{0,5,10\}$. 
As expected, these histograms deviate more as the test location $l$ increases, although the amount of deviation depends on the FRIQ metric.
For PSNR, we see the histogram shifting slightly to the right, while for SSIM, the histogram starts to shrink in width.

\Figref{coverage_v_sliceloc_e2ecnf} suggests that one could select a more conservative $\alpha$ to ensure sufficiently high coverage under small distributional shifts, but at the cost of more conservative bounds. 
In fact, this is largely the mechanism behind distributionally robust CP extensions like \citet{Tibshirani:NIPS:19,Barber:AS:23,Cauchois:JASA:24}.
We leave such generalizations to future work. 

\color{black}


\section{Conclusion} \label{sec:conclusion}
For imaging inverse problems, we used conformal prediction to construct bounds on the FRIQ of a recovered image relative to the unknown true image.
When constructed using a calibration set that is statistically exchangeable with the test sample, our bounds are guaranteed to hold with high probability.
Two of our methods leveraged approximate-posterior-sampling schemes to yield tighter conformal bounds that adapt to the measurements and reconstruction. 
Our approaches were demonstrated on image denoising and accelerated multicoil MRI, illustrating the broad applicability of our work. 

\subsubsection*{Limitations} 
We acknowledge multiple limitations in our proposed methodology.
1) Our methods require access to calibration data $\{(x_i,y_i)\}_{i=1}^n$ that is similar enough to the test data $(x_0,y_0)$ for the FRIQ pairs $\{(z_i,\hat{z}_i)\}_{i=0}^n$ to be modeled as statistically exchangeable.
More work is required to make our methods robust to distribution shift (see \secref{exchange}), although \citet{Tibshirani:NIPS:19,Barber:AS:23,Cauchois:JASA:24} suggest some paths forward.
2) Our methods will be most impactful when there exists evidence that the FRIQ metric is well matched to the application (e.g., DISTS for MRI \citep{Kastryulin:IA:23}).
For some applications, additional work is required to determine which metrics are more appropriate.
3) Our MRI application ideas are preliminary and not ready for practical use; rigorous clinical trials are needed to tune and validate the methodology on a much larger and diverse cohort of data.
4) The learned adaptive bound from \secref{improved} requires training a quantile regression model, and our FFHQ denoising experiment suggests that it may not be easy to significantly outperform the simpler adaptive bound from \secref{adaptive}. 
5) The posterior samplers that we considered in our numerical experiments target only aleatoric uncertainty, and sharper conformal bounds might be attained with posterior samplers that also target epistemic uncertainty (e.g., \citet{Ekmekci:NIPSW:23}).
6) Because our methods are based on CP (or, equivalently, conformal risk control under the indicator loss \citep{Angelopoulos:22}), the marginal guarantee \eqref{friq_coverage} holds with probability $1-\alpha$ over random test data (e.g., $\hat{Z}_0, Z_0$) and calibration sets $D\cal$. 
A more fine-grained coverage could be achieved via the Risk-Controlling Prediction Sets (RCPS) framework from \citet{Bates:JACM:21}, which employs \emph{two} user-selected error rates $\alpha,\delta\in(0,1)$ to yield coverage guarantees like
\begin{align}
\Pr\big[ 
\Pr\big\{
Z_0 \in \mc{C}_{\hat{\lambda}(D\cal)}\big(\hat{Z}_0 \big)
\big| D\cal \big\} \geq 1-\alpha
\big] \geq 1-\delta
\label{eq:adaptive_coverage_rcps}
\end{align}
in place of \eqref{friq_coverage}.
In \eqref{adaptive_coverage_rcps}, $\alpha$ controls the $D\cal$-conditional error while $\delta$ controls the error over $D\cal$.

\subsubsection*{Broader Impact Statement}
By providing conformal bounds on the FRIQ of recovered images, we anticipate that our framework will positively impact the field of imaging inverse problems by providing rigorous guarantees on recovery accuracy. 
Our methods may help to give confidence that recovered images can be trusted, especially in safety-critical applications. 
However, the marginal coverage guarantee in \eqref{friq_coverage} holds on average across random calibration and test data, but not conditionally for a given measurement $y_0$ and/or calibration set $d\cal$.
Furthermore, before our methods are ready for practical use, clinical studies with large and diverse datasets must be performed to guide the choice of FRIQ metrics, error rates $\alpha$, recovery schemes $h(\cdot)$, and conformal bounding strategies.

\section*{Acknowledgements}

This work was supported in part by the National Institutes of Health under Grant R01-EB029957.

\clearpage
\bibliography{bibs/macros,bibs/machine,bibs/misc,bibs/mri,bibs/sparse,bibs/books,bibs/phase}

\begin{thebibliography}{78}
\providecommand{\natexlab}[1]{#1}
\providecommand{\url}[1]{\texttt{#1}}
\expandafter\ifx\csname urlstyle\endcsname\relax
  \providecommand{\doi}[1]{doi: #1}\else
  \providecommand{\doi}{doi: \begingroup \urlstyle{rm}\Url}\fi

\bibitem[Adler \& {\"O}ktem(2018)Adler and {\"O}ktem]{Adler:18}
Jonas Adler and Ozan {\"O}ktem.
\newblock Deep {B}ayesian inversion.
\newblock \emph{arXiv:1811.05910}, 2018.

\bibitem[Andersen et~al.(2023)Andersen, Dahl, and Vandenberghe]{cvxopt}
Martin~S. Andersen, Joachim Dahl, and Lieven Vandenberghe.
\newblock {CVXOPT}, 2023.
\newblock URL \url{https://cvxopt.org/copyright.html}.

\bibitem[Angelopoulos \& Bates(2023)Angelopoulos and
  Bates]{Angelopoulos:FTML:23}
Anastasios~N. Angelopoulos and Stephen Bates.
\newblock Conformal prediction: {A} gentle introduction.
\newblock \emph{Foundations and Trends in Machine Learning}, 16\penalty0
  (4):\penalty0 494--591, 2023.
\newblock ISSN 1935-8237.
\newblock \doi{10.1561/2200000101}.

\bibitem[Angelopoulos et~al.(2022{\natexlab{a}})Angelopoulos, Bates, Fisch,
  Lei, and Schuster]{Angelopoulos:22}
Anastasios~N Angelopoulos, Stephen Bates, Adam Fisch, Lihua Lei, and Tal
  Schuster.
\newblock Conformal risk control.
\newblock \emph{arXiv:2208.02814}, 2022{\natexlab{a}}.

\bibitem[Angelopoulos et~al.(2022{\natexlab{b}})Angelopoulos, Kohli, Bates,
  Jordan, Malik, Alshaabi, Upadhyayula, and Romano]{Angelopoulos:ICML:22}
Anastasios~N. Angelopoulos, Amit~P. Kohli, Stephen Bates, Michael~I. Jordan,
  Jitendra Malik, Thayer Alshaabi, Srigokul Upadhyayula, and Yaniv Romano.
\newblock Image-to-image regression with distribution-free uncertainty
  quantification and applications in imaging.
\newblock In \emph{Proc. Intl. Conf. on Machine Learning}, 2022{\natexlab{b}}.
\newblock \doi{10.48550/arXiv.2202.05265}.

\bibitem[Arridge et~al.(2019)Arridge, Maass, {\"O}ktem, and
  Sch{\"o}nlieb]{Arridge:AN:19}
Simon Arridge, Peter Maass, Ozan {\"O}ktem, and Carola-Bibiane Sch{\"o}nlieb.
\newblock Solving inverse problems using data-driven models.
\newblock \emph{Acta Numerica}, 28:\penalty0 1--174, June 2019.

\bibitem[Banerji et~al.(2023)Banerji, Chakraborti, Harbron, and
  MacArthur]{Banerji:NM:23}
Christopher R.~S. Banerji, Tapabrata Chakraborti, Chris Harbron, and Ben~D.
  MacArthur.
\newblock Clinical {AI} tools must convey predictive uncertainty for each
  individual patient.
\newblock \emph{Nature Medicine}, 29\penalty0 (12):\penalty0 2996--2998, 2023.
\newblock \doi{10.1038/s41591-023-02562-7}.

\bibitem[Barbano et~al.(2021)Barbano, Zhang, Arridge, and Jin]{Barbano:ICPR:21}
Riccardo Barbano, Chen Zhang, Simon Arridge, and Bangti Jin.
\newblock Quantifying model uncertainty in inverse problems via {B}ayesian deep
  gradient descent.
\newblock In \emph{Proc. IEEE Intl. Conf. on Pattern Recognition}, pp.\
  1392--1399, 2021.
\newblock \doi{10.1109/ICPR48806.2021.9412521}.

\bibitem[Barber et~al.(2023)Barber, Candes, Ramdas, and
  Tibshirani]{Barber:AS:23}
Rina~Foygel Barber, Emmanuel~J Candes, Aaditya Ramdas, and Ryan~J Tibshirani.
\newblock Conformal prediction beyond exchangeability.
\newblock \emph{Annals of Statistics}, 51\penalty0 (2):\penalty0 816--845,
  2023.

\bibitem[Bates et~al.(2021)Bates, Angelopoulos, Lei, Malik, and
  Jordan]{Bates:JACM:21}
Stephen Bates, Anastasios Angelopoulos, Lihua Lei, Jitendra Malik, and Michael
  Jordan.
\newblock Distribution-free, risk-controlling prediction sets.
\newblock \emph{Journal of the ACM}, 68\penalty0 (6), 2021.
\newblock \doi{10.1145/3478535}.

\bibitem[Belhasin et~al.(2023)Belhasin, Romano, Freedman, Rivlin, and
  Elad]{Belhasin:TPAMI:23}
Omer Belhasin, Yaniv Romano, Daniel Freedman, Ehud Rivlin, and Michael Elad.
\newblock Principal uncertainty quantification with spatial correlation for
  image restoration problems.
\newblock \emph{IEEE Trans. on Pattern Analysis and Machine Intelligence},
  46:\penalty0 3321--3333, 2023.

\bibitem[Belthangady \& Royer(2019)Belthangady and Royer]{Belthangady:NMe:19}
Chinmay Belthangady and Loic~A Royer.
\newblock Applications, promises, and pitfalls of deep learning for
  fluorescence image reconstruction.
\newblock \emph{Nature Methods}, 16\penalty0 (12):\penalty0 1215--1225, 2019.

\bibitem[Bendel et~al.(2023)Bendel, Ahmad, and Schniter]{Bendel:NIPS:23}
Matthew Bendel, Rizwan Ahmad, and Philip Schniter.
\newblock A regularized conditional {GAN} for posterior sampling in inverse
  problems.
\newblock In \emph{Proc. Neural Information Processing Systems Conf.}, 2023.

\bibitem[Bhadra et~al.(2021)Bhadra, Kelkar, Brooks, and
  Anastasio]{Bhadra:TMI:21}
Sayantan Bhadra, Varun~A Kelkar, Frank~J Brooks, and Mark~A Anastasio.
\newblock On hallucinations in tomographic image reconstruction.
\newblock \emph{IEEE Trans. on Medical Imaging}, 40\penalty0 (11):\penalty0
  3249--3260, 2021.

\bibitem[Blau \& Michaeli(2018)Blau and Michaeli]{Blau:CVPR:18}
Yochai Blau and Tomer Michaeli.
\newblock The perception-distortion tradeoff.
\newblock In \emph{Proc. IEEE Conf. on Computer Vision and Pattern
  Recognition}, pp.\  6228--6237, 2018.

\bibitem[Borovec et~al.(2022)Borovec, Schock, Harsh, Koker, Liello, Stancl,
  Quan, Grechkin, and Falcon]{torchmetric}
Nicki Skafte Detlefsenand~Jiri Borovec, Justus Schock, Ananya Harsh, Teddy
  Koker, Luca~Di Liello, Daniel Stancl, Changsheng Quan, Maxim Grechkin, and
  William Falcon.
\newblock Torchmetrics, 2022.
\newblock URL \url{https://github.com/Lightning-AI/torchmetrics}.

\bibitem[Caron et~al.(2024)Caron, Arnström, Bonagiri, Dechaume, Flowers,
  Heins, Ishikawa, Kenefake, Mazzamuto, Meoli, O'Donoghue, Oppenheimer,
  Pandala, Quiroz~Omaña, Rontsis, Shah, St-Jean, Vitucci, Wolfers, and
  Yang]{qpsolvers}
Stéphane Caron, Daniel Arnström, Suraj Bonagiri, Antoine Dechaume, Nikolai
  Flowers, Adam Heins, Takuma Ishikawa, Dustin Kenefake, Giacomo Mazzamuto,
  Donato Meoli, Brendan O'Donoghue, Adam~A. Oppenheimer, Abhishek Pandala,
  Juan~José Quiroz~Omaña, Nikitas Rontsis, Paarth Shah, Samuel St-Jean,
  Nicola Vitucci, Soeren Wolfers, and Fengyu Yang.
\newblock qpsolvers: Quadratic programming solvers in python, 2024.
\newblock URL \url{https://github.com/qpsolvers/qpsolvers}.

\bibitem[Cauchois et~al.(2024)Cauchois, Gupta, Ali, and
  Duchi]{Cauchois:JASA:24}
Maxime Cauchois, Suyash Gupta, Alnur Ali, and John~C Duchi.
\newblock Robust validation: {C}onfident predictions even when distributions
  shift.
\newblock \emph{Journal of the American Statistical Association}, pp.\  1--66,
  2024.

\bibitem[Chen \& Guestrin(2016)Chen and Guestrin]{Chen:ICKDDM:16}
Tianqi Chen and Carlos Guestrin.
\newblock {XGBoost: A} scalable tree boosting system.
\newblock In \emph{Proc. Intl. Conf. on Knowledge Discovery and Data Mining},
  pp.\  785--794, 2016.

\bibitem[Chu et~al.(2020)Chu, Anandkumar, Shin, and Fishman]{Chu:JACR:20}
Linda~C Chu, Anima Anandkumar, Hoo~Chang Shin, and Elliot~K Fishman.
\newblock The potential dangers of artificial intelligence for radiology and
  radiologists.
\newblock \emph{Journal of the American College of Radiology}, 17\penalty0
  (10):\penalty0 1309--1311, 2020.

\bibitem[Cohen et~al.(2018)Cohen, Luck, and Honari]{Cohen:MICCAI:18}
Joseph~Paul Cohen, Margaux Luck, and Sina Honari.
\newblock Distribution matching losses can hallucinate features in medical
  image translation.
\newblock In \emph{Proc. Intl. Conf. on Medical Image Computation and
  Computer-Assisted Intervention}, pp.\  529--536, 2018.

\bibitem[Ding et~al.(2020{\natexlab{a}})Ding, Ma, Wang, and
  Simoncelli]{Ding:TPAMI:20}
Keyan Ding, Kede Ma, Shiqi Wang, and Eero~P Simoncelli.
\newblock Image quality assessment: {U}nifying structure and texture
  similarity.
\newblock \emph{IEEE Trans. on Pattern Analysis and Machine Intelligence},
  44\penalty0 (5):\penalty0 2567--2581, 2020{\natexlab{a}}.

\bibitem[Ding et~al.(2020{\natexlab{b}})Ding, Ma, Wang, and
  Simoncelli]{Ding:github:20}
Keyan Ding, Kede Ma, Shiqi Wang, and Eero~P. Simoncelli.
\newblock {DISTS}, 2020{\natexlab{b}}.
\newblock URL \url{https://github.com/dingkeyan93/DISTS}.

\bibitem[Durmus et~al.(2018)Durmus, Moulines, and Pereyra]{Durmus:JIS:18}
Alain Durmus, Eric Moulines, and Marcelo Pereyra.
\newblock Efficient {B}ayesian computation by proximal {M}arkov chain {M}onte
  {C}arlo: {W}hen {L}angevin meets {M}oreau.
\newblock \emph{SIAM Journal on Imaging Sciences}, 11\penalty0 (1):\penalty0
  473--506, 2018.

\bibitem[Edupuganti et~al.(2021)Edupuganti, Mardani, Vasanawala, and
  Pauly]{Edupuganti:TMI:20}
Vineet Edupuganti, Morteza Mardani, Shreyas Vasanawala, and John Pauly.
\newblock Uncertainty quantification in deep {MRI} reconstruction.
\newblock \emph{IEEE Trans. on Medical Imaging}, 40\penalty0 (1):\penalty0
  239--250, January 2021.

\bibitem[Ekmekci \& Cetin(2022)Ekmekci and Cetin]{Ekmekci:TCI:22}
Canberk Ekmekci and Mujdat Cetin.
\newblock Uncertainty quantification for deep unrolling-based computational
  imaging.
\newblock \emph{IEEE Trans. on Computational Imaging}, 8:\penalty0 1195--1209,
  2022.
\newblock \doi{10.1109/TCI.2022.3233185}.

\bibitem[Ekmekci \& Cetin(2023)Ekmekci and Cetin]{Ekmekci:NIPSW:23}
Canberk Ekmekci and Mujdat Cetin.
\newblock Quantifying generative model uncertainty in posterior sampling
  methods for computational imaging.
\newblock In \emph{Proc. Neural Information Processing Systems Workshop}, 2023.

\bibitem[Falcon et~al.(2019)]{lightning}
William Falcon et~al.
\newblock Pytorch lightning, 2019.
\newblock URL \url{https://github.com/PyTorchLightning/pytorch-lightning}.

\bibitem[Fristedt \& Gray(2013)Fristedt and Gray]{Fristedt:Book:13}
Bert~E Fristedt and Lawrence~F Gray.
\newblock \emph{A Modern Approach to Probability Theory}.
\newblock Springer, 2013.

\bibitem[Gottschling et~al.(2023)Gottschling, Antun, Hansen, and
  Adcock]{Gottschling:23}
Nina~M Gottschling, Vegard Antun, Anders~C Hansen, and Ben Adcock.
\newblock The troublesome kernel---{O}n hallucinations, no free lunches and the
  accuracy-stability trade-off in inverse problems.
\newblock \emph{arXiv:2001.01258}, 2023.

\bibitem[Hammernik et~al.(2023)Hammernik, K{\"u}stner, Yaman, Huang, Rueckert,
  Knoll, and Ak{\c{c}}akaya]{Hammernik:SPM:23}
Kerstin Hammernik, Thomas K{\"u}stner, Burhaneddin Yaman, Zhengnan Huang,
  Daniel Rueckert, Florian Knoll, and Mehmet Ak{\c{c}}akaya.
\newblock Physics-driven deep learning for computational magnetic resonance
  imaging: {C}ombining physics and machine learning for improved medical
  imaging.
\newblock \emph{IEEE Signal Processing Magazine}, 40\penalty0 (1):\penalty0
  98--114, 2023.

\bibitem[Ho et~al.(2020)Ho, Jain, and Abbeel]{Ho:NIPS:20}
Jonathan Ho, Ajay Jain, and Pieter Abbeel.
\newblock Denoising diffusion probabilistic models.
\newblock In \emph{Proc. Neural Information Processing Systems Conf.},
  volume~33, pp.\  6840--6851, 2020.

\bibitem[Hoffman et~al.(2021)Hoffman, Slavitt, and Fitzpatrick]{Hoffman:NMe:21}
David~P Hoffman, Isaac Slavitt, and Casey~A Fitzpatrick.
\newblock The promise and peril of deep learning in microscopy.
\newblock \emph{Nature Methods}, 18\penalty0 (2):\penalty0 131--132, 2021.

\bibitem[Horwitz \& Hoshen(2022)Horwitz and Hoshen]{Horwitz:22}
Eliahu Horwitz and Yedid Hoshen.
\newblock Conffusion: {C}onfidence intervals for diffusion models.
\newblock \emph{arXiv.2211.09795}, 2022.
\newblock \doi{10.48550/arXiv.2211.09795}.

\bibitem[Jalal et~al.(2021)Jalal, Arvinte, Daras, Price, Dimakis, and
  Tamir]{Jalal:NIPS:21}
Ajil Jalal, Marius Arvinte, Giannis Daras, Eric Price, Alex Dimakis, and
  Jonathan Tamir.
\newblock Robust compressed sensing {MRI} with deep generative priors.
\newblock In \emph{Proc. Neural Information Processing Systems Conf.}, 2021.

\bibitem[Joshi et~al.(2022)Joshi, Pruitt, Chen, Liu, and Ahmad]{Joshi:22}
Mihir Joshi, Aaron Pruitt, Chong Chen, Yingmin Liu, and Rizwan Ahmad.
\newblock Technical report (v1.0)--pseudo-random cartesian sampling for dynamic
  {MRI}.
\newblock \emph{arXiv:2206.03630}, 2022.

\bibitem[Karras et~al.(2018)Karras, Aila, Laine, and Lehtinen]{Karras:ICLR:18}
Tero Karras, Timo Aila, Samuli Laine, and Jaakko Lehtinen.
\newblock Progressive growing of {GAN}s for improved quality, stability, and
  variation.
\newblock In \emph{Proc. Intl. Conf. on Learning Representations}, 2018.

\bibitem[Karras et~al.(2019)Karras, Laine, and Aila]{Karras:CVPR:19}
Tero Karras, Samuli Laine, and Timo Aila.
\newblock A style-based generator architecture for generative adversarial
  networks.
\newblock In \emph{Proc. IEEE Conf. on Computer Vision and Pattern
  Recognition}, pp.\  4396--4405, 2019.

\bibitem[Kastryulin et~al.(2023)Kastryulin, Zakirov, Pezzotti, and
  Dylov]{Kastryulin:IA:23}
Sergey Kastryulin, Jamil Zakirov, Nicola Pezzotti, and Dmitry~V Dylov.
\newblock Image quality assessment for magnetic resonance imaging.
\newblock \emph{IEEE Access}, 11:\penalty0 14154--14168, 2023.

\bibitem[Kawar et~al.(2022{\natexlab{a}})Kawar, Elad, Ermon, and
  Song]{Kawar:NIPS:22}
Bahjat Kawar, Michael Elad, Stefano Ermon, and Jiaming Song.
\newblock Denoising diffusion restoration models.
\newblock In \emph{Proc. Neural Information Processing Systems Conf.},
  2022{\natexlab{a}}.

\bibitem[Kawar et~al.(2022{\natexlab{b}})Kawar, Elad, Ermon, and
  Song]{Kawar:github:22}
Bahjat Kawar, Michael Elad, Stefano Ermon, and Jiaming Song.
\newblock Denoising diffusion restoration models.
\newblock Downloaded from \url{https://github.com/bahjat-kawar/ddrm}, May
  2022{\natexlab{b}}.

\bibitem[Kendall \& Gal(2017)Kendall and Gal]{Kendall:NIPS:17}
Alex Kendall and Yarin Gal.
\newblock What uncertainties do we need in {B}ayesian deep learning for
  computer vision?
\newblock In \emph{Proc. Neural Information Processing Systems Conf.}, 2017.

\bibitem[Knoll et~al.(2020)Knoll, Hammernik, Zhang, Moeller, Pock, Sodickson,
  and Akcakaya]{Knoll:SPM:20}
Florian Knoll, Kerstin Hammernik, Chi Zhang, Steen Moeller, Thomas Pock,
  Daniel~K Sodickson, and Mehmet Akcakaya.
\newblock Deep-learning methods for parallel magnetic resonance imaging
  reconstruction: {A} survey of the current approaches, trends, and issues.
\newblock \emph{IEEE Signal Processing Magazine}, 37\penalty0 (1):\penalty0
  128--140, January 2020.

\bibitem[Koenker \& Bassett(1978)Koenker and Bassett]{Koenker:ECON:78}
Roger Koenker and Gilbert Bassett.
\newblock Regression quantiles.
\newblock \emph{Econometrica}, 46\penalty0 (1), 1978.
\newblock \doi{10.2307/1913643}.

\bibitem[Kutiel et~al.(2023)Kutiel, Cohen, Elad, Freedman, and
  Rivlin]{Kutiel:ICLR:23}
Gilad Kutiel, Regev Cohen, Michael Elad, Daniel Freedman, and Ehud Rivlin.
\newblock Conformal prediction masks: {V}isualizing uncertainty in medical
  imaging.
\newblock In \emph{Proc. Intl. Conf. on Learning Representations}, 2023.

\bibitem[Laumont et~al.(2022)Laumont, Bortoli, Almansa, Delon, Durmus, and
  Pereyra]{Laumont:JIS:22}
R{\'e}mi Laumont, Valentin~De Bortoli, Andr{\'e}s Almansa, Julie Delon, Alain
  Durmus, and Marcelo Pereyra.
\newblock Bayesian imaging using plug \& play priors: {W}hen {L}angevin meets
  {T}weedie.
\newblock \emph{SIAM Journal on Imaging Sciences}, 15\penalty0 (2):\penalty0
  701--737, 2022.

\bibitem[Lei \& Wasserman(2014)Lei and Wasserman]{Lei:JRSS:14}
Jing Lei and Larry Wasserman.
\newblock Distribution-free prediction bands for non-parametric regression.
\newblock \emph{Journal of the Royal Statistical Society}, 76, 2014.
\newblock \doi{10.1111/rssb.12021}.

\bibitem[Lei et~al.(2018)Lei, G'Sell, Rinaldo, Tibshirani, and
  Wasserman]{Lei:JASA:18}
Jing Lei, Max G'Sell, Alessandro Rinaldo, Ryan~J. Tibshirani, and Larry
  Wasserman.
\newblock Distribution-free predictive inference for regression.
\newblock \emph{Journal of the American Statistical Association}, 2018.

\bibitem[Lin \& Kuo(2011)Lin and Kuo]{Lin:JVCIR:11}
Weisi Lin and C-C~Jay Kuo.
\newblock Perceptual visual quality metrics: {A} survey.
\newblock \emph{Journal of Visual Communication and Image Representation},
  22\penalty0 (4):\penalty0 297--312, 2011.

\bibitem[Lindsay(2021)]{Lindsay:JCNS:21}
Grace~W Lindsay.
\newblock Convolutional neural networks as a model of the visual system:
  {P}ast, present, and future.
\newblock \emph{Journal of Cognitive Neuroscience}, 33\penalty0 (10):\penalty0
  2017--2031, 2021.

\bibitem[Muckley et~al.(2021)Muckley, Riemenschneider, Radmanesh, Kim, Jeong,
  Ko, Jun, Shin, Hwang, Mostapha, et~al.]{Muckley:TMI:21}
Matthew~J Muckley, Bruno Riemenschneider, Alireza Radmanesh, Sunwoo Kim, Geunu
  Jeong, Jingyu Ko, Yohan Jun, Hyungseob Shin, Dosik Hwang, Mahmoud Mostapha,
  et~al.
\newblock Results of the 2020 {fastMRI} challenge for machine learning {MR}
  image reconstruction.
\newblock \emph{IEEE Trans. on Medical Imaging}, 40\penalty0 (9):\penalty0
  2306--2317, 2021.

\bibitem[Narnhofer et~al.(2022)Narnhofer, Effland, Kobler, Hammernik, Knoll,
  and Pock]{Narnhofer:TMI:22}
Dominik Narnhofer, Alexander Effland, Erich Kobler, Kerstin Hammernik, Florian
  Knoll, and Thomas Pock.
\newblock Bayesian uncertainty estimation of learned variational {MRI}
  reconstruction.
\newblock \emph{IEEE Trans. on Medical Imaging}, 41\penalty0 (2):\penalty0
  279--291, 2022.

\bibitem[Narnhofer et~al.(2024)Narnhofer, Habring, Holler, and
  Pock]{Narnhofer:JIS:24}
Dominik Narnhofer, Andreas Habring, Martin Holler, and Thomas Pock.
\newblock Posterior-variance-based error quantification for inverse problems in
  imaging.
\newblock \emph{SIAM Journal on Imaging Sciences}, 17\penalty0 (1):\penalty0
  301--333, 2024.

\bibitem[Papadopoulos et~al.(2002)Papadopoulos, Proedrou, Vovk, and
  Gammerman]{Papadopoulos:ECML:02}
Harris Papadopoulos, Kostas Proedrou, Volodya Vovk, and Alex Gammerman.
\newblock Inductive confidence machines for regression.
\newblock In \emph{Proc. European Conf. on Machine Learning}, pp.\  345--356,
  2002.
\newblock \doi{10.1007/3-540-36755-1_29}.

\bibitem[Paszke et~al.(2019)Paszke, Gross, Massa, Lerer, Bradbury, Chanan,
  Killeen, Lin, Gimelshein, Antiga, Desmaison, Kopf, Yang, DeVito, Raison,
  Tejani, Chilamkurthy, Steiner, Fang, Bai, and Chintala]{pytorch}
Adam Paszke, Sam Gross, Francisco Massa, Adam Lerer, James Bradbury, Gregory
  Chanan, Trevor Killeen, Zeming Lin, Natalia Gimelshein, Luca Antiga, Alban
  Desmaison, Andreas Kopf, Edward Yang, Zachary DeVito, Martin Raison, Alykhan
  Tejani, Sasank Chilamkurthy, Benoit Steiner, Lu~Fang, Junjie Bai, and Soumith
  Chintala.
\newblock {PyTorch: A}n imperative style, high-performance deep learning
  library.
\newblock In \emph{Proc. Neural Information Processing Systems Conf.}, pp.\
  8024--8035, 2019.

\bibitem[Roemer et~al.(1990)Roemer, Edelstein, Hayes, Souza, and
  Mueller]{Roemer:MRM:90}
Peter~B Roemer, William~A Edelstein, Cecil~E Hayes, Steven~P Souza, and
  Otward~M Mueller.
\newblock The {NMR} phased array.
\newblock \emph{Magnetic Resonance in Medicine}, 16\penalty0 (2):\penalty0
  192--225, 1990.

\bibitem[Romano et~al.(2019)Romano, Patterson, and Cand{\`e}s]{Romano:NIPS:19}
Yaniv Romano, Evan Patterson, and Emmanuel~J. Cand{\`e}s.
\newblock Conformalized quantile regression.
\newblock In \emph{Proc. Neural Information Processing Systems Conf.}, pp.\
  3543--3553, 2019.
\newblock \doi{10.48550/arXiv.1905.03222}.

\bibitem[Sankaranarayanan et~al.(2022)Sankaranarayanan, Angelopoulos, Bates,
  Romano, and Isola]{Sankaranarayanan:NIPS:22}
Swami Sankaranarayanan, Anastasios~N. Angelopoulos, Stephen Bates, Yaniv
  Romano, and Phillip Isola.
\newblock Semantic uncertainty intervals for disentangled latent spaces.
\newblock In \emph{Proc. Neural Information Processing Systems Conf.}, 2022.
\newblock \doi{10.48550/arXiv.2207.10074}.

\bibitem[Sriram et~al.(2020{\natexlab{a}})Sriram, Zbontar, Murrell, Defazio,
  Zitnick, Yakubova, Knoll, and Johnson]{Sriram:MICCAI:20}
Anuroop Sriram, Jure Zbontar, Tullie Murrell, Aaron Defazio, C.~Lawrence
  Zitnick, Nafissa Yakubova, Florian Knoll, and Patricia Johnson.
\newblock End-to-end variational networks for accelerated {MRI} reconstruction.
\newblock In \emph{Proc. Intl. Conf. on Medical Image Computation and
  Computer-Assisted Intervention}, pp.\  64--73, 2020{\natexlab{a}}.

\bibitem[Sriram et~al.(2020{\natexlab{b}})Sriram, Zbontar, Murrell, Defazio,
  Zitnick, Yakubova, Knoll, and Johnson]{Sriram:github:20}
Anuroop Sriram, Jure Zbontar, Tullie Murrell, Aaron Defazio, C.~Lawrence
  Zitnick, Nafissa Yakubova, Florian Knoll, and Patricia Johnson.
\newblock End-to-end variational networks for accelerated {MRI} reconstruction.
\newblock \url{https://github.com/facebookresearch/fastMRI},
  2020{\natexlab{b}}.

\bibitem[Sukthanker et~al.(2022)Sukthanker, Huang, Kumar, Timofte, and
  Van~Gool]{Sukthanker:CVPR:22}
Rhea~Sanjay Sukthanker, Zhiwu Huang, Suryansh Kumar, Radu Timofte, and Luc
  Van~Gool.
\newblock Generative flows with invertible attentions.
\newblock In \emph{Proc. IEEE Conf. on Computer Vision and Pattern
  Recognition}, 2022.
\newblock \doi{10.48550/arXiv.2106.03959}.

\bibitem[Sun \& Bouman(2021)Sun and Bouman]{Sun:AAAI:21}
He~Sun and Katherine~L Bouman.
\newblock Deep probabilistic imaging: {U}ncertainty quantification and
  multi-modal solution characterization for computational imaging.
\newblock In \emph{Proc. AAAI Conf. Artificial Intelligence}, volume~35, pp.\
  2628--2637, 2021.

\bibitem[Teneggi et~al.(2023)Teneggi, Tivnan, Stayman, and
  Sulam]{Teneggi:ICML:23}
Jacopo Teneggi, Matthew Tivnan, J.~Webster Stayman, and Jeremias Sulam.
\newblock How to trust your diffusion model: {A} convex optimization approach
  to conformal risk control, 2023.

\bibitem[Tibshirani et~al.(2019)Tibshirani, Foygel~Barber, Candes, and
  Ramdas]{Tibshirani:NIPS:19}
Ryan~J Tibshirani, Rina Foygel~Barber, Emmanuel Candes, and Aaditya Ramdas.
\newblock Conformal prediction under covariate shift.
\newblock In \emph{Proc. Neural Information Processing Systems Conf.},
  volume~32, 2019.

\bibitem[Tivnan et~al.(2024)Tivnan, Yoon, Chen, Li, Wu, and
  Li]{Tivnan:MICCAI:24}
Matthew Tivnan, Siyeop Yoon, Zhennong Chen, Xiang Li, Dufan Wu, and Quanzheng
  Li.
\newblock Hallucination index: {A}n image quality metric for generative
  reconstruction models.
\newblock In \emph{Proc. Intl. Conf. on Medical Image Computation and
  Computer-Assisted Intervention}, pp.\  449--458, 2024.

\bibitem[Tonolini et~al.(2020)Tonolini, Radford, Turpin, Faccio, and
  Murray-Smith]{Tonolini:JMLR:20}
Francesco Tonolini, Jack Radford, Alex Turpin, Daniele Faccio, and Roderick
  Murray-Smith.
\newblock Variational inference for computational imaging inverse problems.
\newblock \emph{Journal of Machine Learning Research}, 21\penalty0
  (179):\penalty0 1--46, 2020.

\bibitem[Vovk et~al.(2005)Vovk, Gammerman, and Shafer]{Vovk:Book:05}
Vladimir Vovk, Alexander Gammerman, and Glenn Shafer.
\newblock \emph{Algorithmic Learning in a Random World}.
\newblock Springer, New York, 2005.

\bibitem[Wang(2011)]{Wang:SPM:11}
Zhou Wang.
\newblock Applications of objective image quality assessment methods.
\newblock \emph{IEEE Signal Processing Magazine}, 28\penalty0 (6):\penalty0
  137--142, 2011.

\bibitem[Wang et~al.(2004)Wang, Bovik, Sheikh, and Simoncelli]{Wang:TIP:04}
Zhou Wang, Alan~C Bovik, Hamid~R Sheikh, and Eero~P Simoncelli.
\newblock Image quality assessment: {F}rom error visibility to structural
  similarity.
\newblock \emph{IEEE Trans. on Image Processing}, 13\penalty0 (4):\penalty0
  600--612, April 2004.

\bibitem[Wen et~al.(2023{\natexlab{a}})Wen, Ahmad, and Schniter]{Wen:ICML:23}
Jeffrey Wen, Rizwan Ahmad, and Philip Schniter.
\newblock A conditional normalizing flow for accelerated multi-coil {MR}
  imaging.
\newblock In \emph{Proc. Intl. Conf. on Machine Learning}, 2023{\natexlab{a}}.

\bibitem[Wen et~al.(2023{\natexlab{b}})Wen, Ahmad, and Schniter]{Wen:github:23}
Jeffrey Wen, Rizwan Ahmad, and Philip Schniter.
\newblock {MRI CNF}, 2023{\natexlab{b}}.
\newblock URL \url{https://github.com/jwen307/mri_cnf}.

\bibitem[Wen et~al.(2024)Wen, Ahmad, and Schniter]{Wen:24}
Jeffrey Wen, Rizwan Ahmad, and Philip Schniter.
\newblock Task-driven uncertainty quantification in inverse problems via
  conformal prediction.
\newblock \emph{arXiv:2405.18527}, 2024.

\bibitem[Wu et~al.(2024)Wu, Sun, Chen, Zhang, Yue, and Bouman]{Wu:24}
Zihui Wu, Yu~Sun, Yifan Chen, Bingliang Zhang, Yisong Yue, and Katherine~L
  Bouman.
\newblock Principled probabilistic imaging using diffusion models as
  plug-and-play priors.
\newblock \emph{arXiv:2405.18782}, 2024.

\bibitem[Xue et~al.(2019)Xue, Cheng, Li, and Tian]{Xue:Optica:19}
Yujia Xue, Shiyi Cheng, Yunzhe Li, and Lei Tian.
\newblock Reliable deep-learning-based phase imaging with uncertainty
  quantification.
\newblock \emph{Optica}, 6\penalty0 (5), 2019.
\newblock ISSN 2334-2536.
\newblock \doi{10.1364/OPTICA.6.000618}.

\bibitem[Yamins \& DiCarlo(2016)Yamins and DiCarlo]{Yamins:NNS:16}
Daniel~LK Yamins and James~J DiCarlo.
\newblock Using goal-driven deep learning models to understand sensory cortex.
\newblock \emph{Nature neuroscience}, 19\penalty0 (3):\penalty0 356--365, 2016.

\bibitem[Zach et~al.(2023)Zach, Knoll, and Pock]{Zach:TMI:23}
Martin Zach, Florian Knoll, and Thomas Pock.
\newblock Stable deep {MRI} reconstruction using generative priors.
\newblock \emph{IEEE Trans. on Medical Imaging}, 2023.

\bibitem[Zbontar et~al.(2018)Zbontar, Knoll, Sriram, Muckley, Bruno, Defazio,
  Parente, Geras, Katsnelson, Chandarana, Zhang, Drozdzal, Romero, Rabbat,
  Vincent, Pinkerton, Wang, Yakubova, Owens, Zitnick, Recht, Sodickson, and
  Lui]{Zbontar:18}
Jure Zbontar, Florian Knoll, Anuroop Sriram, Matthew~J. Muckley, Mary Bruno,
  Aaron Defazio, Marc Parente, Krzysztof~J. Geras, Joe Katsnelson, Hersh
  Chandarana, Zizhao Zhang, Michal Drozdzal, Adriana Romero, Michael Rabbat,
  Pascal Vincent, James Pinkerton, Duo Wang, Nafissa Yakubova, Erich Owens,
  C.~Lawrence Zitnick, Michael~P. Recht, Daniel~K. Sodickson, and Yvonne~W.
  Lui.
\newblock fast{MRI: An} open dataset and benchmarks for accelerated {MRI}.
\newblock \emph{arXiv:1811.08839}, 2018.

\bibitem[Zhang et~al.(2018)Zhang, Isola, Efros, Shechtman, and
  Wang]{Zhang:CVPR:18}
Richard Zhang, Phillip Isola, Alexei~A Efros, Eli Shechtman, and Oliver Wang.
\newblock The unreasonable effectiveness of deep features as a perceptual
  metric.
\newblock In \emph{Proc. IEEE Conf. on Computer Vision and Pattern
  Recognition}, pp.\  586--595, 2018.

\end{thebibliography}
\bibliographystyle{tmlr}
\clearpage

\appendix

\section{Empirical coverage} \label{app:empirical_coverage}
In \secref{denoising}, we empirically demonstrated that the coverage guarantees in \eqref{friq_coverage} are met for the non-adaptive, quantile, and regression bounds in the FFHQ denoising experiments. 
Here, we further demonstrate that these guarantees hold regardless of the number of posterior samples $c$ used to compute the adaptive bounds.
Tables \ref{tab:quantile_coverage} and \ref{tab:regression_coverage} show the average empirical coverage for the quantile and regression method, respectively, across $T=10\,000$ trials for different values of $c$ and $\alpha=0.05$.
The same number of posterior samples $c$ is used during calibration and to compute the adaptive bounds during testing.
Again, we observe that the average empirical coverage is very close to the desired $1-\alpha$ in all cases though there are very slight deviations as a result of finite trials, number of calibration samples, and number of testing samples.

\begin{table}[t]
    \centering
    \caption{Mean empirical coverage for the quantile method with $\alpha=0.05$ and $T=10\,000$ on the FFHQ denoising task ($\pm$ standard error)}
    \label{tab:quantile_coverage}
    \begin{tabular}{ccccc}
        \toprule
        $c$ & DISTS & LPIPS & PSNR & SSIM \\
        \midrule
        1  & $0.95002 \pm 0.00009$ & $0.94997 \pm 0.00009$ & $0.95013 \pm 0.00009$ & $0.94989 \pm 0.00009$ \\
        2  & $0.95006 \pm 0.00009$ & $0.95003 \pm 0.00009$ & $0.95001 \pm 0.00009$ & $0.95022 \pm 0.00009$ \\
        4  & $0.94997 \pm 0.00009$ & $0.95008 \pm 0.00009$ & $0.94986 \pm 0.00009$ & $0.94999 \pm 0.00009$ \\
        8  & $0.95020 \pm 0.00009$ & $0.95015 \pm 0.00009$ & $0.95019 \pm 0.00009$ & $0.94991 \pm 0.00009$ \\
        16 & $0.94998 \pm 0.00009$ & $0.94999 \pm 0.00009$ & $0.95009 \pm 0.00009$ & $0.95008 \pm 0.00009$ \\
        32 & $0.95002 \pm 0.00009$ & $0.95013 \pm 0.00009$ & $0.95003 \pm 0.00009$ & $0.95006 \pm 0.00009$ \\
        \bottomrule
    \end{tabular}
\end{table}

\begin{table}[t!]
    \centering
    \caption{Mean empirical coverage for the regression method with $\alpha=0.05$ and $T=10\,000$ on the FFHQ denoising task ($\pm$ standard error)}
    \label{tab:regression_coverage}
    \begin{tabular}{ccccc}
        \toprule
        $c$ & DISTS & LPIPS & PSNR & SSIM \\
        \midrule
        1  & $0.94994 \pm 0.00009$ & $0.94970 \pm 0.00009$ & $0.95009 \pm 0.00009$ & $0.95014 \pm 0.00009$ \\
        2  & $0.95011 \pm 0.00009$ & $0.94953 \pm 0.00009$ & $0.94985 \pm 0.00009$ & $0.95004 \pm 0.00009$ \\
        4  & $0.94996 \pm 0.00009$ & $0.94946 \pm 0.00009$ & $0.95003 \pm 0.00009$ & $0.94995 \pm 0.00009$ \\
        8  & $0.95004 \pm 0.00009$ & $0.94964 \pm 0.00009$ & $0.94999 \pm 0.00009$ & $0.95017 \pm 0.00009$ \\
        16 & $0.94986 \pm 0.00009$ & $0.94964 \pm 0.00009$ & $0.95007 \pm 0.00009$ & $0.94987 \pm 0.00009$ \\
        32 & $0.95013 \pm 0.00009$ & $0.95026 \pm 0.00009$ & $0.95001 \pm 0.00009$ & $0.95006 \pm 0.00009$ \\
        \bottomrule
    \end{tabular}
\end{table}

\begin{table}[t]
    \centering
    \captionof{table}{Mean empirical coverage for the quantile method across accelerations with $\alpha=0.05$, $c=32$, and $T=10\,000$ on the accelerated MRI task ($\pm$ standard error). All coverages are above the expected coverage of $1-\alpha=0.95$}
    \label{tab:mri_coverage_e2ecnf}
    \begin{tabular}{ccccc}
        \hline
        $R$ & DISTS & LPIPS & PSNR & SSIM \\
        \hline
        2  & $0.9503 \pm 0.0001$ & $0.9503 \pm 0.0001$ & $0.9504 \pm 0.0001$ & $0.9504 \pm 0.0001$ \\
        4  & $0.9504 \pm 0.0001$ & $0.9503 \pm 0.0001$ & $0.9505 \pm 0.0001$ & $0.9504 \pm 0.0001$ \\
        8  & $0.9503 \pm 0.0001$ & $0.9504 \pm 0.0001$ & $0.9503 \pm 0.0001$ & $0.9503 \pm 0.0001$ \\
        16 & $0.9504 \pm 0.0001$ & $0.9504 \pm 0.0001$ & $0.9505 \pm 0.0001$ & $0.9506 \pm 0.0001$ \\
        \hline
    \end{tabular}
\end{table}

In \Tabref{mri_coverage_e2ecnf}, we report the mean empirical coverage for the quantile method in the MRI experiments with $\alpha=0.05$, $c=32$, and acceleration rate $R \in \{2,4,8,16\}$ across $T=10\,000$ trials.
For any value of $R$, we see the empirical coverage is very close to the theoretical $1-\alpha=0.95$ coverage; thus, once again, our method shows close compliance to the theory.

\section{Additional FFHQ denoising experiments} \label{app:additiona_ffhq}
\begin{figure}[t]
    \centering
    \includegraphics[width=1\linewidth]{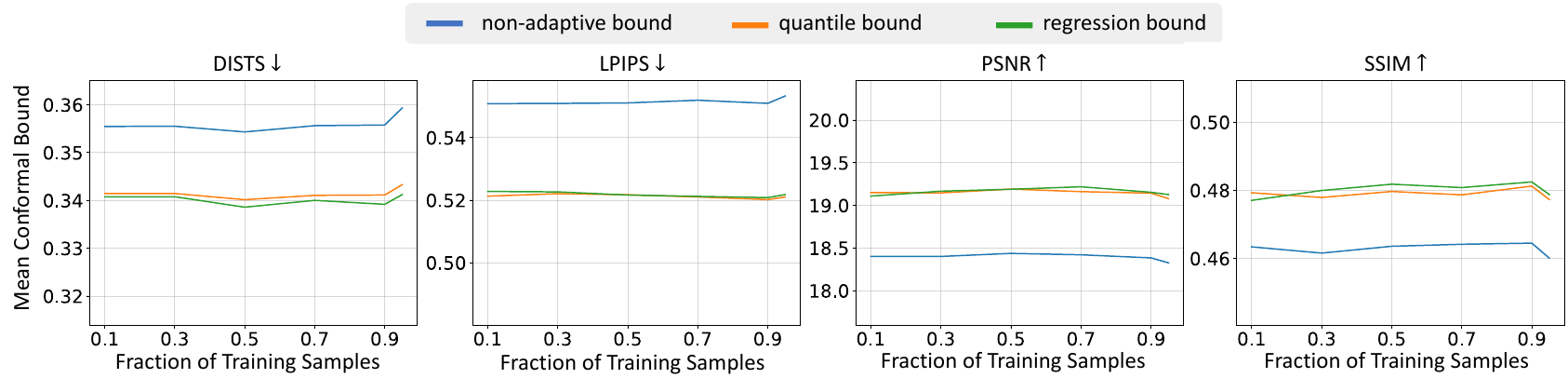}
    \caption{Mean conformal bound versus the proportion of training samples for FFHQ denoising with $n\train + n\cal=3100$ samples.}
    \label{fig:ddrm_bound_v_training_samples}
\end{figure}

\textbf{Effect of training and calibration set size:}~
For FFHQ denoising, we now investigate how the amount of training and calibration data affect the mean conformal bound. 
Following the same Monte Carlo procedure as \secref{denoising}, we fix the number of testing samples to $900$ but change the proportion of $n\train$ versus $n\cal$ for the remaining $3100$ samples. 
\textb{For the non-adaptive and quantile bounds, the training samples are unused.}
In \figref{ddrm_bound_v_training_samples}, we show the mean conformal bounds as the proportion of training samples varies, starting with 0.1 and going up to $0.95$, for $T=10\,000$, $c=32$, and $\alpha=0.05$. 
Both adaptive methods still provide noticeable gains over the non-adaptive bound.
Even with additional training samples, however, the regression bounds show relatively little improvement over the quantile bounds.
Based on \eqref{lambda_hat}, the conformal bounds should grow more conservative as the number of calibration points decreases for the non-adaptive and quantile bounds.
However, this effect is not evident until very small calibration set sizes (e.g., when the fraction of calibration samples is $0.05$).

\begin{figure}[t!]
    \centering
    \includegraphics[width=1\linewidth]{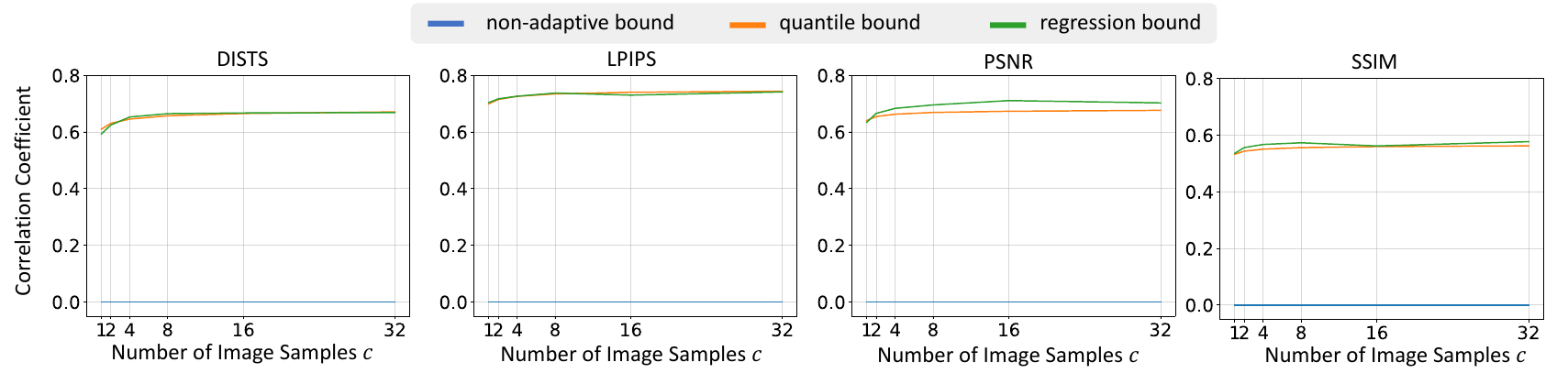}
    \caption{Mean Pearson correlation coefficient between each conformal bound and the true FRIQ versus the number of posterior samples $c$ for FFHQ denoising.}
    \label{fig:corr_v_c}
\end{figure}

\textbf{Correlation between conformal bound and true FRIQ:}~
\Figref{ddrm_qualitative} visually demonstrates that the quantile bound tracks the true FRIQ much better than the non-adaptive bound. 
To quantify this tracking behavior, we compute the Pearson correlation coefficient between each conformal bound $\beta(\hat{z}_k, \hat{\lambda}(d\cal[t]))$ and the true FRIQ $z_k$ over the test indices $k \in \mc{I}\test[t]$ for each Monte-Carlo trial $t$.
In \figref{corr_v_c}, we plot the mean (across $T=10000$ trials) Pearson correlation coefficient versus $c$ for each bound. 
Since the non-adaptive bound is constant with $z_k$, its correlation equals 0.
However, the two adaptive approaches demonstrate a correlation coefficient above $0.5$, and up to $0.7$, depending on the metric.
These correlation coefficients quantify the adaptivity of our bounds and explain, in part, why the adaptive bounds led to better average acceleration rates than the non-adaptive bound in the multi-round measurement experiment of \secref{MRI}.

\begin{figure}[t]
    \centering
    \begin{minipage}{0.48\linewidth}
        \centering
        \includegraphics[width=1.0\linewidth,trim= 0 0 0 0,clip]{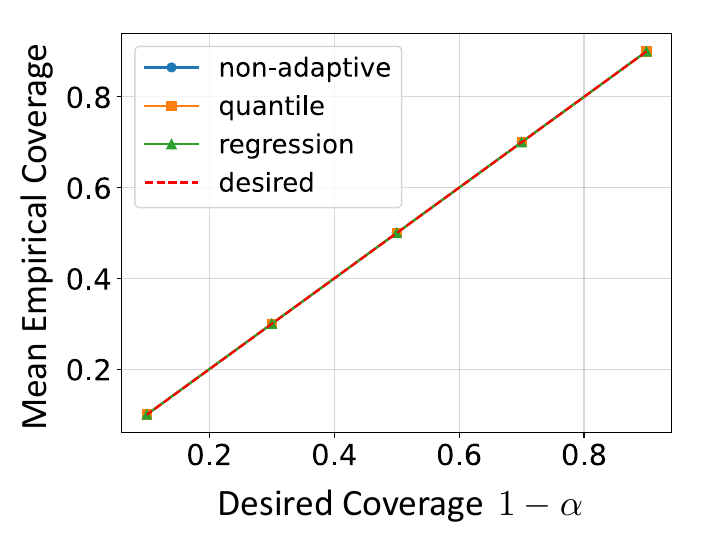}
        \caption{\textb{The average empirical coverage across $T=10000$ trials for different values of $1-\alpha$. Results are shown for LPIPS with $c=32$.}}
        \label{fig:ddrm_cov_vs_alpha}
    \end{minipage}
    \hfill
    \begin{minipage}{0.48\linewidth}
        \centering
        \includegraphics[width=1.0\linewidth,trim= 0 10 0 0,clip]{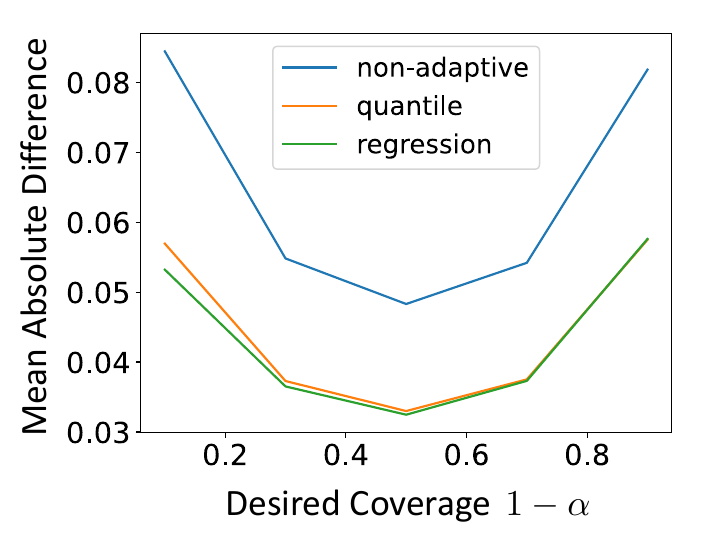}
        \caption{\textb{The mean absolute difference between the conformal bound and true image quality for varying $1-\alpha$. Results are shown for LPIPS with $c=32$. }
        }
        \label{fig:ddrm_mean_abs_diff_vs_alpha}
    \end{minipage}
\end{figure}

\textb{
\textbf{Performance across different error rates $\alpha$:}~
In \figref{ddrm_cov_vs_alpha}, we plot the mean empirical coverage for different values of $1-\alpha$ across $T=10000$ trials. 
Results are shown for each bound on the LPIPS metric with $c=32$. 
As expected, the empirical coverage for all three bounds is very close to the desired $1-\alpha$ for all values of $\alpha$, further demonstrating adherence to conformal prediction theory.
For the same experimental setup, we plot the mean absolute difference between the conformal bound and true image quality against the desired coverage $1-\alpha$ in \figref{ddrm_mean_abs_diff_vs_alpha}. 
Once again, we see the quantile and regression bounds give tighter bounds on average compared to the non-adaptive bounds.
The tightest bounds for all three methods are found when $\alpha=0.5$ since the conditional median is known to be the minimum mean absolute error (MMAE) estimator.
}

\begin{figure}[t]
    \centering
    \includegraphics[width=1\linewidth]{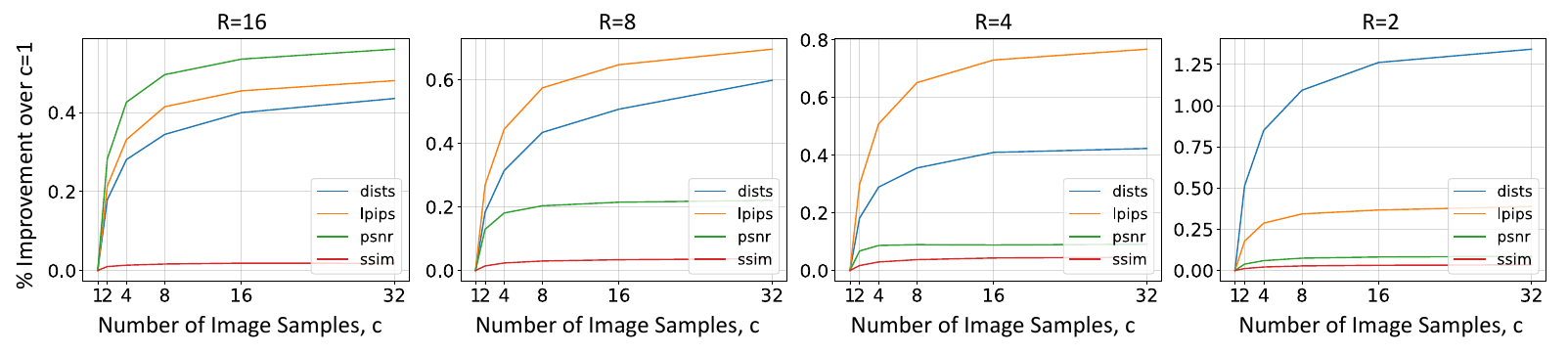}
    \caption{Percent improvement in MCB versus number of samples $c$ used in the quantile bound for the accelerated MRI experiments.}
    \label{fig:mri_bound_v_c}
\end{figure}

\begin{figure}[t]
    \centering
    \includegraphics[width=1\linewidth]{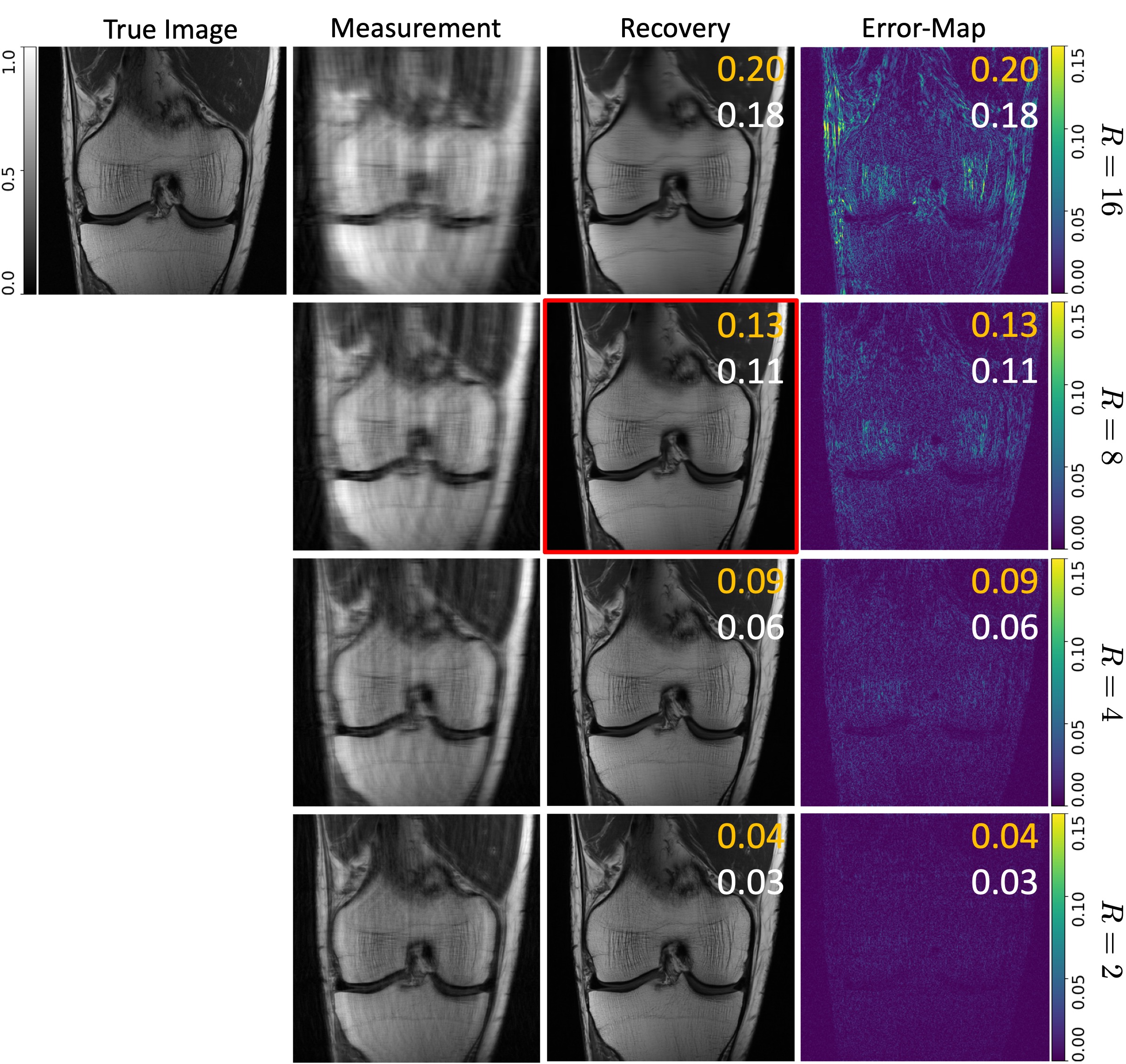}
    \caption{Qualitative example of the multi-round MRI experiment with DISTS at $\alpha=0.05$, $\tau=0.16$, and $c=32$. The measurement, recovery, and absolute error are shown for all accelerations. The quantile bound (orange) and true DISTS (white) are imposed on the reconstructions. The red box indicates the accepted reconstruction where the bound first falls below the threshold $\tau$.}
    \label{fig:mri_qualitative_all}
\end{figure}

\section{Additional MRI experiments} \label{app:add_mri_experiments}
\textbf{Effect of number of posterior samples $c$ in conformal bound:}~
For the case of FFHQ denoising, \secref{denoising} demonstrated the number of posterior samples $c$ has a limited effect on the conformal bounds for the FFHQ experiments.
We now investigate whether the same occurs with MRI.
\Figref{mri_bound_v_c} plots the percent improvement in MCB as $c$ increases relative to the MCB for $c=1$.
From the figure, we see less than a $1.5\%$ improvement over $c=1$ for any metric, suggesting that the quantile method is indeed robust to the choice of $c$ for both experiments.

\textbf{Multi-round measurement samples:}
In \figref{mri_qualitative_all}, we show the zero-filled measurement, recovered image, and absolute-error map at each acceleration rate.
The conformal bound is imposed on the reconstructions for the case when $\alpha=0.05$, $\tau=0.16$, and $c=32$.
Following the multi-round measurement protocol described in \secref{MRI}, the reconstruction at $R=8$ (marked in red) would be deemed sufficient ($\beta_i < \tau$), and the measurement collection would end.

\section{MRI reconstruction from a CNF} \label{app:cnf}


In the MRI experiments of \secref{MRI}, we used the E2E-VarNet from \citet{Sriram:MICCAI:20} for MRI image recovery and the CNF from \citet{Wen:ICML:23} adaptive bound computation.
In this section, we investigate what happens when the CNF from \citet{Wen:ICML:23} is also used for image recovery.

When using an approximate posterior sampler for image recovery, one has the option of averaging $p\geq 1$ posterior samples.
\color{black}
%
%
For example, when one is interested in estimating $x_i$ from $y_i$ with high PSNR, or equivalently low MSE, it makes sense to set $\hat{x}_i$ as the minimum MSE (MMSE) or conditional-mean estimate $\E\{X_i|Y_i\!=\!y_i\}$.
The MMSE estimate can be approximated by the empirical mean 
\begin{align}
\hat{x}_i
= \frac{1}{p} \sum_{j=c+1}^{c+p} \tilde{x}_i\of{j} 
\label{eq:avg},
\end{align}
with large $p$.
The indices on $j$ in \eqref{avg} are chosen to avoid the samples $\{\tilde{x}_i\of{j}\}_{j=1}^c$ used for the adaptive bounds.

However, because the MMSE estimate can look unrealistically smooth, one may instead be interested in producing an image estimate with good SSIM, DISTS, or LPIPS performance.
In this case, smaller values of $p$ may be more appropriate.
For example, \citet{Bendel:NIPS:23} found that, for multicoil brain MRI at acceleration $R=8$ with their conditional GAN, the best choices were $p=8$ for SSIM and $p=2$ for both DISTS and LPIPS.
This is can explained by the perception-distortion tradeoff \citep{Blau:CVPR:18}, which says that, as the MSE distortion improves, the perceptual quality tends to decrease. 
In the end, each FRIQ metric prefers a particular tradeoff between perceptual quality and distortion, and thus a particular $p$ when sample averaging.
In the sequel, we consider different choices for $p$ when using the CNF for image recovery.

\textbf{Average reconstruction performance:}~
To get a sense for how the CNF compares to the E2E-VarNet, and for how the choice of $p$ affects the CNF, we report average PSNR, SSIM, DISTS, and LPIPS on the non-fat-suppressed subset of the fastMRI knee validation set at acceleration rates $R=$ 16, 8, 4, and 2 in Tables~\ref{tab:mri_recon_r16}, \ref{tab:mri_recon_r8}, \ref{tab:mri_recon_r4}, and \ref{tab:mri_recon_r2}, respectively. 
There we see that the E2E-VarNet outperforms the CNF in PSNR and SSIM at all accelerations, while the CNF outperforms the E2E-VarNet in DISTS and LPIPS in all cases other than LPIPS at $R=2$. 

\begin{table}[t]
\centering
\caption{Average reconstruction performance on the fastMRI \citep{Zbontar:18} knee validation set for $R=16$ ($\pm$ standard error)}
\label{tab:mri_recon_r16}
\begin{tabular}{c c c c c}
\toprule
Network & DISTS $\downarrow$ & LPIPS $\downarrow$ & PSNR $\uparrow$ & SSIM $\uparrow$\\
\midrule
E2E-VarNet & $0.209 \pm 0.001$ & $0.354 \pm 0.001$ & $\mathbf{30.301 \pm 0.043}$ & $\mathbf{0.807 \pm 0.001}$ \\
CNF ($p=1$) & $0.183 \pm 0.001$ & $0.312 \pm 0.001$ & $28.244 \pm 0.039$ & $0.688 \pm 0.002$ \\
CNF ($p=2$) & $0.167 \pm 0.001$ & $0.292 \pm 0.001$ & $29.091 \pm 0.039$ & $0.730 \pm 0.001$ \\
CNF ($p=4$) & $\mathbf{0.165 \pm 0.001}$ & $\mathbf{0.287 \pm 0.001}$ & $29.588 \pm 0.039$ & $0.755 \pm 0.001$ \\
CNF ($p=8$) & $0.173 \pm 0.001$ & $0.296 \pm 0.001$ & $29.862 \pm 0.039$ & $0.770 \pm 0.001$ \\
CNF ($p=16$) & $0.184 \pm 0.001$ & $0.314 \pm 0.001$ & $30.006 \pm 0.039$ & $0.777 \pm 0.001$ \\
CNF ($p=32$) & $0.193 \pm 0.001$ & $0.333 \pm 0.001$ & $30.080 \pm 0.039$ & $0.781 \pm 0.001$ \\
\bottomrule
\end{tabular}
\end{table}

\begin{table}[t!]
\centering
\caption{Average reconstruction performance on the fastMRI \citep{Zbontar:18} knee validation set for $R=8$ ($\pm$ standard error)}
\label{tab:mri_recon_r8}
\begin{tabular}{c c c c c}
\toprule
Network & DISTS $\downarrow$ & LPIPS $\downarrow$ & PSNR $\uparrow$ & SSIM $\uparrow$\\
\midrule
E2E-VarNet & $0.151 \pm 0.001$ & $0.262 \pm 0.001$ & $\mathbf{33.459 \pm 0.047}$ & $\mathbf{0.864 \pm 0.001}$ \\
CNF ($p=1$)  & $0.136 \pm 0.000$ & $0.248 \pm 0.001$ & $30.796 \pm 0.044$ & $0.761 \pm 0.002$ \\
CNF ($p=2$) & $\mathbf{0.118 \pm 0.000}$ & $0.225 \pm 0.001$ & $31.754 \pm 0.044$ & $0.799 \pm 0.001$ \\
CNF ($p=4$) & $0.119 \pm 0.000$ & $\mathbf{0.219 \pm 0.001}$ & $32.329 \pm 0.043$ & $0.821 \pm 0.001$ \\
CNF ($p=8$)  & $0.128 \pm 0.001$ & $0.228 \pm 0.001$ & $32.650 \pm 0.043$ & $0.834 \pm 0.001$ \\
CNF ($p=16$) & $0.138 \pm 0.001$ & $0.243 \pm 0.001$ & $32.819 \pm 0.043$ & $0.840 \pm 0.001$ \\
CNF ($p=32$) & $0.145 \pm 0.001$ & $0.255 \pm 0.001$ & $32.907 \pm 0.043$ & $0.843 \pm 0.001$ \\
\bottomrule
\end{tabular}
\end{table}

\begin{table}[t!]
\centering
\caption{Average reconstruction performance on the fastMRI \citep{Zbontar:18} knee validation set for $R=4$ ($\pm$ standard error)}
\label{tab:mri_recon_r4}
\begin{tabular}{c c c c c}
\toprule
Network & DISTS $\downarrow$ & LPIPS $\downarrow$ & PSNR $\uparrow$ & SSIM $\uparrow$ \\
\midrule
E2E-VarNet & $0.110 \pm 0.001$ & $0.181 \pm 0.001$ & $\mathbf{36.030 \pm 0.053}$ & $\mathbf{0.905 \pm 0.001}$ \\
CNF ($p=1$) & $0.100 \pm 0.000$ & $0.191 \pm 0.001$ & $33.090 \pm 0.048$ & $0.826 \pm 0.001$ \\
CNF ($p=2$) & $\mathbf{0.087 \pm 0.000}$ & $0.170 \pm 0.001$ & $34.073 \pm 0.048$ & $0.856 \pm 0.001$ \\
CNF ($p=4$) & $0.090 \pm 0.000$ & $\mathbf{0.166 \pm 0.001}$ & $34.666 \pm 0.048$ & $0.873 \pm 0.001$ \\
CNF ($p=8$) & $0.099 \pm 0.000$ & $0.171 \pm 0.001$ & $34.998 \pm 0.047$ & $0.882 \pm 0.001$ \\
CNF ($p=16$) & $0.106 \pm 0.001$ & $0.178 \pm 0.001$ & $35.174 \pm 0.047$ & $0.887 \pm 0.001$ \\
CNF ($p=32$) & $0.110 \pm 0.001$ & $0.184 \pm 0.001$ & $35.265 \pm 0.047$ & $0.889 \pm 0.001$ \\
\bottomrule
\end{tabular}
\end{table}

\begin{table}[t!]
\centering
\caption{Average reconstruction performance on the fastMRI \citep{Zbontar:18} knee validation set for $R=2$ ($\pm$ standard error)}
\label{tab:mri_recon_r2}
\begin{tabular}{c c c c c}
\toprule
Network & DISTS $\downarrow$ & LPIPS $\downarrow$ & PSNR $\uparrow$ & SSIM $\uparrow$ \\
\midrule
E2E-VarNet & $0.059 \pm 0.000$ & $\mathbf{0.094 \pm 0.001}$ & $\mathbf{39.692 \pm 0.060}$ & $\mathbf{0.947 \pm 0.001}$ \\
CNF ($p=1$) & $0.059 \pm 0.000$ & $0.118 \pm 0.000$ & $36.810 \pm 0.054$ & $0.907 \pm 0.001$ \\
CNF ($p=2$) & $\mathbf{0.054 \pm 0.000}$ & $0.105 \pm 0.000$ & $37.667 \pm 0.054$ & $0.923 \pm 0.001$ \\
CNF ($p=4$) & $0.055 \pm 0.000$ & $0.100 \pm 0.000$ & $38.171 \pm 0.054$ & $0.931 \pm 0.001$ \\
CNF ($p=8$) & $0.058 \pm 0.000$ & $0.099 \pm 0.000$ & $38.448 \pm 0.054$ & $0.935 \pm 0.001$ \\
CNF ($p=16$) & $0.060 \pm 0.000$ & $0.099 \pm 0.000$ & $38.593 \pm 0.054$ & $0.937 \pm 0.001$ \\
CNF ($p=32$) & $0.061 \pm 0.000$ & $0.099 \pm 0.000$ & $38.668 \pm 0.054$ & $0.939 \pm 0.001$ \\
\bottomrule
\end{tabular}
\end{table}

\color{black}

\textbf{Empirical Coverage:}~
We first investigate empirical coverage using the same Monte-Carlo validation procedure described in \secref{MRI}, again with $T=10000$ trials.
Table~\ref{tab:mri_coverage} reports the average empirical coverage for the quantile bounds with different choices of $p$ at $R=8$.
As with all previous experiments, the coverage is above and very close to the desired $1-\alpha$ value for all choices of $p$.

\begin{table}[t]
    \centering
    \captionof{table}{Mean empirical coverage for the quantile method with $\alpha=0.05$, $c=32$, and $T=10\,000$ on the $R=8$ accelerated MRI task ($\pm$ standard error). All coverages are above the expected coverage of $1-\alpha=0.95$}
    \label{tab:mri_coverage}
    \begin{tabular}{ccccc}
        \hline
        $p$ & DISTS & LPIPS & PSNR & SSIM \\
        \hline
        1  & $0.9503 \pm 0.0001$ & $0.9503 \pm 0.0001$ & $0.9505 \pm 0.0001$ & $0.9504 \pm 0.0001$ \\
        2  & $0.9505 \pm 0.0001$ & $0.9503 \pm 0.0001$ & $0.9504 \pm 0.0001$ & $0.9505 \pm 0.0001$ \\
        4  & $0.9503 \pm 0.0001$ & $0.9503 \pm 0.0001$ & $0.9505 \pm 0.0001$ & $0.9504 \pm 0.0001$ \\
        8  & $0.9505 \pm 0.0001$ & $0.9504 \pm 0.0001$ & $0.9504 \pm 0.0001$ & $0.9505 \pm 0.0001$ \\
        16 & $0.9505 \pm 0.0001$ & $0.9502 \pm 0.0001$ & $0.9504 \pm 0.0001$ & $0.9504 \pm 0.0001$ \\
        32 & $0.9504 \pm 0.0001$ & $0.9506 \pm 0.0001$ & $0.9504 \pm 0.0001$ & $0.9505 \pm 0.0001$ \\
        \hline
    \end{tabular}
\end{table}

\begin{figure}[t]
    \centering
    \includegraphics[width=1\linewidth]{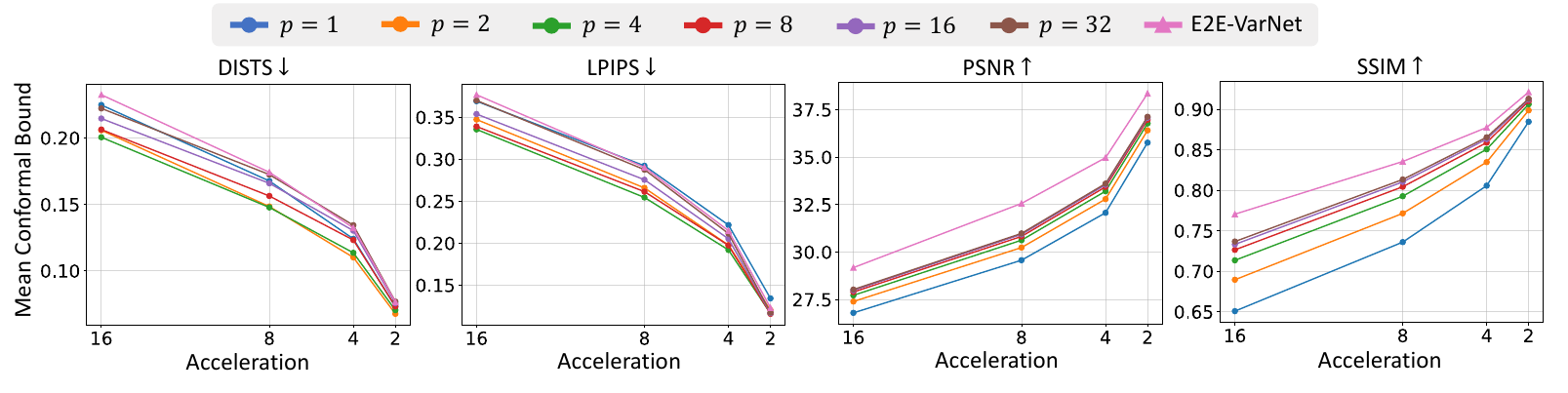}
    \caption{Mean conformal bound versus acceleration $R$ for accelerated MRI.  Results shown for the quantile bound with $\hat{x}$ computed from the E2E-VarNet point estimate (shown in pink) or the $p$-sample average from CNF posteriors. Various $p$ shown.}
    \label{fig:mri_bound_v_accel}
\end{figure}

\textbf{Effect of acceleration rate $R$ and choice of recovery method:}~
\Figref{mri_bound_v_accel} plots the mean conformal bound (MCB) of the quantile method with $c=32$ versus the acceleration rate $R$ for different image estimates $\hat{x}_i$. 
The image estimate $\hat{x}$ is computed using either the E2E-VarNet point estimate or a $p$-sample average from the CNF with different values of $p$. 
In all cases, the MCB improves as the acceleration $R$ decreases, as expected.
But, as discussed above, each metric benefits from a different choice of $p$.
DISTS and LPIPS prefer $p\in\{2,4\}$ while PSNR and SSIM prefer $p=32$. 
The figure also shows that the MCB for the $p$-optimized CNF-based method is better than the MCB for the E2E-VarNet-based method with both DISTS and LPIPS but not with PSNR and SSIM.
Thus, the recovery method that yields the tightest bounds may depend on the metric of interest. 

\begin{figure}[t]
    \centering
    \begin{minipage}{0.45\linewidth}
        \centering
        \includegraphics[width=1.0\linewidth,trim= 0 10 0 0,clip]{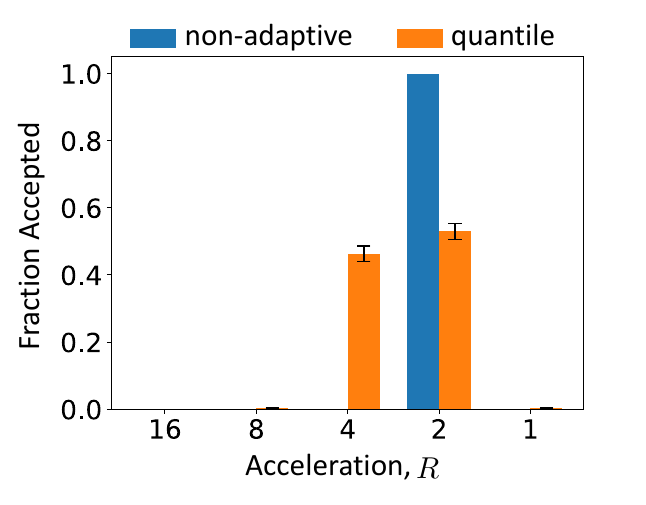}
        \caption{Fraction of accepted slices versus final acceleration rate for multi-round MRI using DISTS. Both methods use a $p$-posterior average for the recovery with $p=4$, $c=32$, $\alpha=0.05$, and $\tau=0.11$. Error bars show standard deviation.
        }
        \label{fig:slices_remaining}
    \end{minipage}
    \hfill
    \begin{minipage}{0.49\linewidth}
        \centering
        \captionof{table}{Average results for a multi-round MRI simulation where measurement collection stop once bounds are below a user-set threshold $\tau$. Results shown for $T=10\,000$ trials using the DISTS metric with $\alpha=0.05$, $\tau=0.11$, $p=4$, and $c=32$ ($\pm$ standard error).
        }
        \resizebox{\columnwidth}{!}{%
        \begin{tabular}{|c|c|c|}
            \hline
            Method 
            & \multicolumn{1}{|p{2cm}|}{\centering Average \\ Acceleration}  
            & \multicolumn{1}{|p{2cm}|}{\centering Acceptance \\ Empirical \\ Coverage} \\
            \hline
            Non-adaptive & $2.000 \pm 0.000$ & $0.9504 \pm 0.0001$  \\
            Quantile & $2.596 \pm 0.001$ & $0.9434 \pm 0.0001$  \\
            \hline
        \end{tabular}
        }
        \label{tab:multi_round}
    \end{minipage}
\end{figure}

\textbf{Multi-round MRI:}~
Since \Figref{mri_bound_v_accel} reveals the tightest bounds on DISTS are obtained when the CNF posterior average with $p=4$ is used for the image estimate $\hat{x}$, we repeat the multi-round experiment from \secref{MRI} with this setup.
As before, we set $\alpha=0.05$ and $c=32$.
The DISTS acceptance threshold is set at $\tau=0.11$, where the non-adaptive approach requires $R=2$ for acceptance. 
In \Figref{slices_remaining}, we plot the number of slices that were collected at each acceleration rate.
Here, the non-adaptive approach always accepts slices at $R=2$ while the quantile bound accepts nearly $50\%$ of the slices at $R=4$.
Table \ref{tab:multi_round} shows that this equates to an average accepted acceleration rate of $2.596$, with an empirical coverage of $0.9434$ at acceptance. 
This demonstrates that, when considering multi-round recovery performance, the relative performance of different bounding strategies may depend on which image recovery model is used. 
In any case, the modularity of our proposed framework allows one to improve multi-round performance through the choice of the recovery model, the posterior sampler, and/or the conformal bound.

\begin{figure}[t]
    \centering
    \begin{minipage}{0.6\linewidth}
        \centering
        \includegraphics[width=1.0\linewidth,trim= 0 0 0 0,clip]{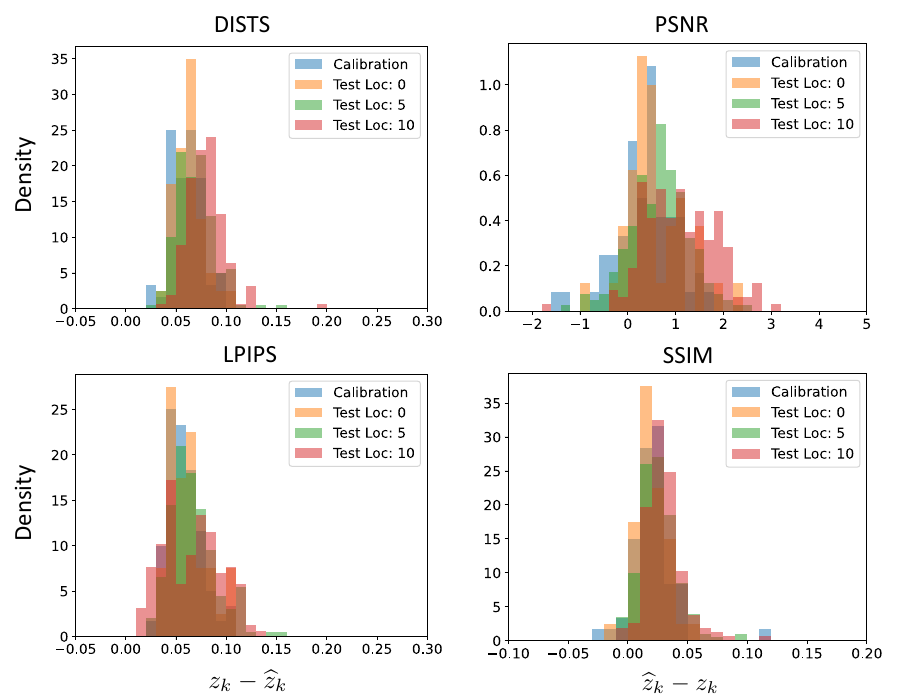}
        \caption{Histograms of the difference between the true FRIQ $z_k$ and the FRIQ estimate $\hat{z}_k$ for test samples $k$ in the test fold of a single trial. Histograms are shown for test slice locations $l=0,5,10$. Note the increasing shift in distribution from the calibration set as $l$ increases.}
        \label{fig:sliceloc_density}
    \end{minipage}
    \hfill
    \begin{minipage}{0.38\linewidth}
        \centering
        \includegraphics[width=1.0\linewidth,trim= 0 10 0 0,clip]{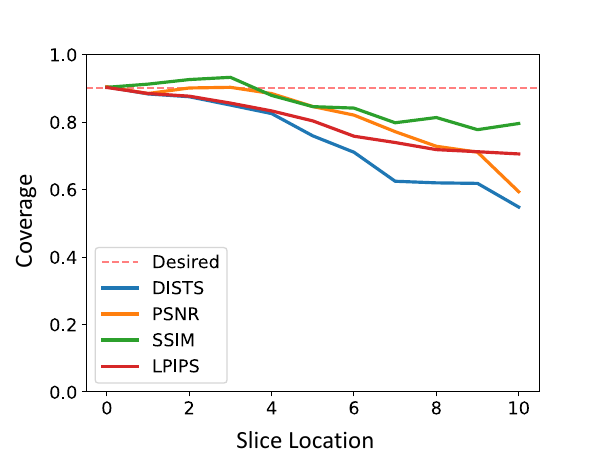}
        \caption{The average empirical coverage across $T=10000$ trials for test sets at different slice locations. All trials are calibrated with images from slice location $0$ with $\alpha=0.1$, $R=8$, $p=1$, and $c=32$. 
        }
        \label{fig:coverage_v_sliceloc}
    \end{minipage}
\end{figure}

\textbf{Distribution Shift:}~
We repeat the distribution shift analysis from \secref{exchange} using a single CNF posterior sample as the image estimate $\hat{x}$, i.e. $p=1$, with $\alpha=0.1$, $c=32$, and $R=8$. 
As before, we can visualize the distribution shift by looking at the histograms of the difference between the true FRIQ $z_k$ and FRIQ estimate $\hat{z}_k$ for each test index $k \in \mc{I}\test[t]$ as the test location $l$ increases. \Figref{sliceloc_density} shows the histograms for test locations $l\in\{0,5,10\}$. 
It shows more distinct distributional shifts compared to \secref{exchange}: as $l$ increases, the PSNR histogram noticeably shifts to the right and widens, while the histogram for LPIPS becomes bimodal.
Not surprisingly, these more dramatic shifts lead to a decrease in coverage for all metrics in \figref{coverage_v_sliceloc} as $l$ increases.
We do note, however, that both SSIM and PSNR retain a coverage at or above $1-\alpha$ until $l=4$, demonstrating a certain level of robustness.

\section{MRI subsampling mask details} \label{app:mri_masks}
For the MRI experiments, we simulate the collection of measurements at four acceleration rates $R=\{16,8,4,2\}$. 
These measurements are collected in the 2D spatial frequency domain known as k-space, and the pattern with which these samples are collected is called a sampling mask. 

\begin{figure}[t]
    \centering
    \includegraphics[width=1\linewidth]{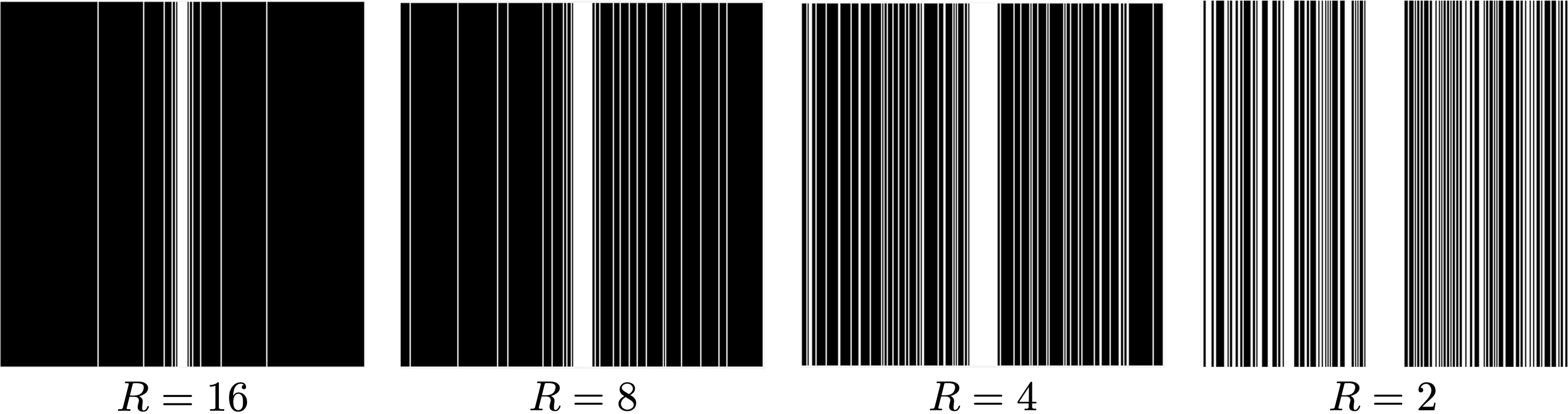}
    \caption{MRI sampling masks in k-space for each acceleration rate $R$. White pixels indicate the measurement was collected for that location in k-space. The masks are designed in a nested fashion where each mask contains all the measurements of higher $R$.}
    \label{fig:sampling_masks}
\end{figure}

For this study, we use a Cartesian sampling procedure where full lines of the 2D k-space are collected progressively.
Starting with $R=16$, we utilize a Golden Ratio Offset (GRO) \citep{Joshi:22} sampling mask with GRO-specific parameters $s=15$ and $\alpha=8$. 
This gives a fully sampled region of 9 lines in the center of k-space known as the autocalibration signal (ACS) region. 
To simulate the iterative collection of measurements, we build upon this mask for $R=8$.
We first collect central lines to obtain an ACS region of 16 lines before sampling additional k-space lines with a sampling probability inversely proportional to the distance from the center.
Additional lines are collected until the desired acceleration $R=8$ is met. 
This procedure is repeated for $R=4$ and $R=2$ to acquire masks with ACS widths of 24 and 32, respectively. 
\figref{sampling_masks} illustrates examples of the resulting masks.

\section{Training/Model details} \label{app:training_details}

For the regression bound in \secref{improved}, where $u_i\in\Real^c$, we use a quantile predictor of the form
\begin{align}
f(u_i;\theta) = \psi( u_i )\tran w + b
~~\text{with}~~ \theta = [w, b]\tran,
\label{eq:spline} ,
\end{align}
where $\psi(\cdot)$ is a linear spline with two knots, $t_1$ and $t_2$, implemented via the truncated power basis
\begin{align}
\psi( u_i ) = [u_i; (u_i-t_1 \underline{1})_{+}; (u_i-t_2 \underline{1})_{+}] \in \Real^{3c},
\end{align}
with $\underline{1}$ the $c$-dimensional vector of ones and $(x)_+\defn\max(x,0)$. 
The two knots were placed at the $\frac{1}{3}$ and $\frac{2}{3}$ empirical quantiles of the mean training feature $\{\frac{1}{c}\sum_{j=1}^{c} \tilde{z}_i\of{j}\}_{i=n\cal+1}^{n\cal+n\train}$, respectively.
Essentially, for each feature in $u_i$, \eqref{spline} implements a piece-wise-linear regression function  with three distinct pieces. 
To promote consistency in $u_i=[\tilde{z}_{i}\of{1},\tilde{z}_{i}\of{2},\dots,\tilde{z}_{i}\of{c}]\tran$ across different $i$, the spline function $\phi(\cdot)$ first sorts the values $\{\tilde{z}_{i}\of{j}\}_{j=1}^c$ within each $u_i$.
For $\rho(\theta)$ in \eqref{QR}, we use ridge regularization on the weights $w$.
The resulting \eqref{QR} is a quadratic program, which can be optimized using any convex solver.
To tune the regularization weight $\gamma$, we use $K$-fold cross validation with $K=5$ folds and select the weight that provides the lowest mean pinball loss across the 5 folds. 

For DDRM, we use the author's implementation \citep{Kawar:github:22}, which is publicly available under an MIT license. 

Both fastMRI reconstruction models were trained once with all four acceleration rates.
For each sample in an epoch, one of the four sampling masks is randomly drawn, allowing the model to see each sample at a different acceleration throughout the training.

With the E2E-VarNet, we use the author's codebase \citep{Sriram:github:20}, which is released under an MIT license.
For training, we utilize the default hyperparameters provided by the authors for the model on the fastMRI knee leaderboard. 
The model was trained for 50 epochs with a batch size of 16 and learning rate of 0.0001 using SSIM \citep{Wang:TIP:04} as the loss function. 
This takes around 38 hours on a single NVIDIA V100 with 32GB of memory.

To train the CNF, we start with the author's implementation \citep{Wen:github:23} that is available under an MIT license. 
We modify the architecture slightly in order to better handle multiple accelerations.
First, we include an invertible attention module, iMAP \citep{Sukthanker:CVPR:22}, to the end of the base flow step. 
Then, we increase the number of initial channels in the conditioning network to 256.
Using 2 layers and 10 flows steps in each layer, we train the CNF to minimize the negative log-likelihood objective.
The model is trained for 150 epochs with batch size 8 and learning rate 0.0001. 
On a single NVIDIA V100, this takes around 335 hours. 

To compute the quadratic program for \secref{denoising}, we use the qpsolver \citep{qpsolvers} package under a LGPL 3.0 license along with the CVXOPT \citep{cvxopt} package under a GNU General Public License. 

We use the TorchMetrics \citep{torchmetric} package under the Apache 2.0 license to compute PSNR, SSIM, and LPIPS.
We use the author's code at \citep{Ding:github:20} for DISTS under a MIT license. 
For multicoil MRI, we first compute the magnitude images using the ``root-sum-of-squares'' (RSS) \citep{Roemer:MRM:90} before computing any metric. 
Since DISTS and LPIPS require a 3-channel image, we repeat the magnitude image for all three channels and normalize the values to be between 0 and 1 before computing either metric. 

All models use the PyTorch \citep{pytorch} framework with a custom license allowing open use.
The E2E-VarNet and CNF are implemented using PyTorch Lightning \citep{lightning} under an Apache 2.0 license. 

\section{Datasets} \label{app:data_licenses}
The Flickr-Faces-HQ (FFHQ) \citep{Karras:CVPR:19} is publicly available under the Creative Commons BY-NC-SA 4.0 license.
The fastMRI \citep{Zbontar:18} datasets is available under a royalty-free license for internal research and educational purposes by the NYU fastMRI initiative. 
The providers have deidentified and manually inspected images and metadata for protected health information (PHI) as part of an IRB-approved study.

\end{document}